\documentclass[journal]{IEEEtran}

\usepackage{ulem}
\usepackage{times}
\usepackage{epsfig}
\usepackage{amsmath}
\usepackage{amssymb}
\usepackage{amsmath}

\usepackage[pagebackref=true,breaklinks=true,colorlinks,bookmarks=false]{hyperref}

\usepackage{enumerate}
\usepackage{multirow}  % 多行合并的宏包
\usepackage{stfloats}
\usepackage{booktabs}
\usepackage{diagbox}
\usepackage{makecell}

\usepackage{graphics}
\usepackage{subfigure}
\graphicspath{{figs/}}

\usepackage{algorithm}
\usepackage{algorithmic}
\usepackage{color}
\usepackage{colortbl}
\usepackage{comment}
\definecolor{LightRed}{rgb}{1,0.92,0.92}
\definecolor{LightOrange}{rgb}{1,0.95,0.88}
\definecolor{LightYellow}{rgb}{1.0,1.0,0.84}
\definecolor{LightGreen}{rgb}{0.9,1.0,0.88}
\definecolor{LightCyan}{rgb}{0.9,1,1}
\definecolor{LightBlue}{rgb}{0.9,0.94,1}
\definecolor{LightIndigo}{rgb}{0.92,0.9,1}
\definecolor{LightMagenta}{rgb}{0.96,0.86,1}
\definecolor{DirtyWhite}{rgb}{0.96,0.96,0.96}

\begin{document}

\newcommand{\revise}[1]{\textcolor{black}{#1}}

%%%%%%%%% TITLE
\title{Vision-Language Models in Remote Sensing: Current Progress and Future Trends}

\author{Xiang Li$^\dagger$$^{,*}$, \textit{IEEE Member}, Congcong Wen$^\dagger$, \textit{IEEE Member}, Yuan Hu$^\dagger$, \\ Zhenghang Yuan, Xiao Xiang Zhu, \textit{IEEE Fellow}.
\thanks{Xiang Li is with the King Abdullah University of Science and Technology, Thuwal 23955, Saudi Arabia (email: xiangli92@ieee.org). Congcong Wen is with the Department of Electrical and Computer Engineering, New York University Abu Dhabi, Abu Dhabi 129188, UAE (email: wencc@nyu.edu). Yuan Hu is with the Institute of Remote Sensing and Geographic Information Systems, Peking University, Beijing 100871, China (email: huyuan@pku.edu.cn). Zhenghang Yuan and Xiao Xiang Zhu are with the Chair of Data Science in Earth Observation, Technical University of Munich, Munich 80333, Germany (email: zhenghang.yuan@tum.de; xiaoxiang.zhu@tum.de). Xiao Xiang Zhu is also with the Munich Center for Machine Learning, Munich 80333, Germany. 
}
\thanks{$*$ Corresponding author: Xiang Li. $\dagger$ Co-first author.}
\thanks{The work of Z. Yuan and X. Zhu is jointly supported by the German Federal Ministry of Education and Research (BMBF) in the framework of the international future AI lab ``AI4EO--Artificial Intelligence for Earth Observation: Reasoning, Uncertainties, Ethics and Beyond" (grant number: 01DD20001) by Munich Center for Machine Learning.}
}

% \markboth{IEEE Transactions on Geoscience and Remote Sensing,~Vol.~13, No.~9, September~2014}%
% {Li \MakeLowercase{\textit{et al.}}: Bare Demo of IEEEtran.cls for Journals}

\maketitle

\begin{abstract}
The remarkable achievements of ChatGPT and GPT-4 have sparked a wave of interest and research in the field of large language models for Artificial General Intelligence (AGI). These models provide intelligent solutions close to human thinking, enabling us to use general artificial intelligence to solve problems in various applications. However, in remote sensing (RS), the scientific literature on the implementation of AGI remains relatively scant. Existing AI-related research in remote sensing primarily focuses on visual understanding tasks while neglecting the semantic understanding of the objects and their relationships. This is where vision-language models excel, as they enable reasoning about images and their associated textual descriptions, allowing for a deeper understanding of the underlying semantics. Vision-language models can go beyond visual recognition of RS images, model semantic relationships, and generate natural language descriptions of the image. This makes them better suited for tasks requiring visual and textual understanding, such as image captioning, and visual question answering. This paper provides a comprehensive review of the research on vision-language models in remote sensing, summarizing the latest progress, highlighting challenges, and identifying potential research opportunities. Specifically, we review the application of vision-language models in mainstream remote sensing tasks, including image captioning, text-based image generation, text-based image retrieval, visual question answering, scene classification, semantic segmentation, and object detection. For each task, we analyze representative works and discuss research progress. We also summarize the commonly used remote sensing vision-language datasets, codebases, and online accessible resources. Finally, we delineate the limitations of current studies and suggest potential avenues for future advancements. This review aims to provide a comprehensive review of the current research progress of vision-language models in remote sensing and to inspire further research in this exciting and promising field.

\end{abstract}


\vspace{-10pt}
\section{Introduction}
\vspace{-5pt}

Deep learning has emerged as a powerful tool for various remote sensing (RS) applications. Early works in RS primarily focused on using visual features extracted from images to perform various tasks, such as object detection, semantic segmentation, and change detection. As one of the most commonly used deep learning methods, convolutional neural networks (CNNs)~\cite{lenet} can automatically learn hierarchical representations of remote sensing images, allowing them to capture local and global spatial features and patterns. Moreover, attention mechanisms~\cite{vaswani2017attention} have been incorporated into deep learning models to improve their performance in RS tasks by allowing the model to focus on specific regions of the input. Thanks to the powerful feature learning abilities of deep neural networks, deep learning models have proven their effectiveness in various RS tasks, achieving state-of-the-art performance compared to traditional machine learning approaches~\cite{zhang2016deep,zhu2017deep}. Nevertheless, existing deep learning-based research mostly focuses on visual understanding tasks while neglecting the semantic understanding of the objects and their relationships. For example, when performing land cover classification, a vision-only model may classify a building rooftop pixel as a highway road if the pixel is visually similar to a highway road. This is because the model lacks the general knowledge that highways can not be inside building rooftops.

\begin{figure*}
    \centering
    \includegraphics[width=15cm]{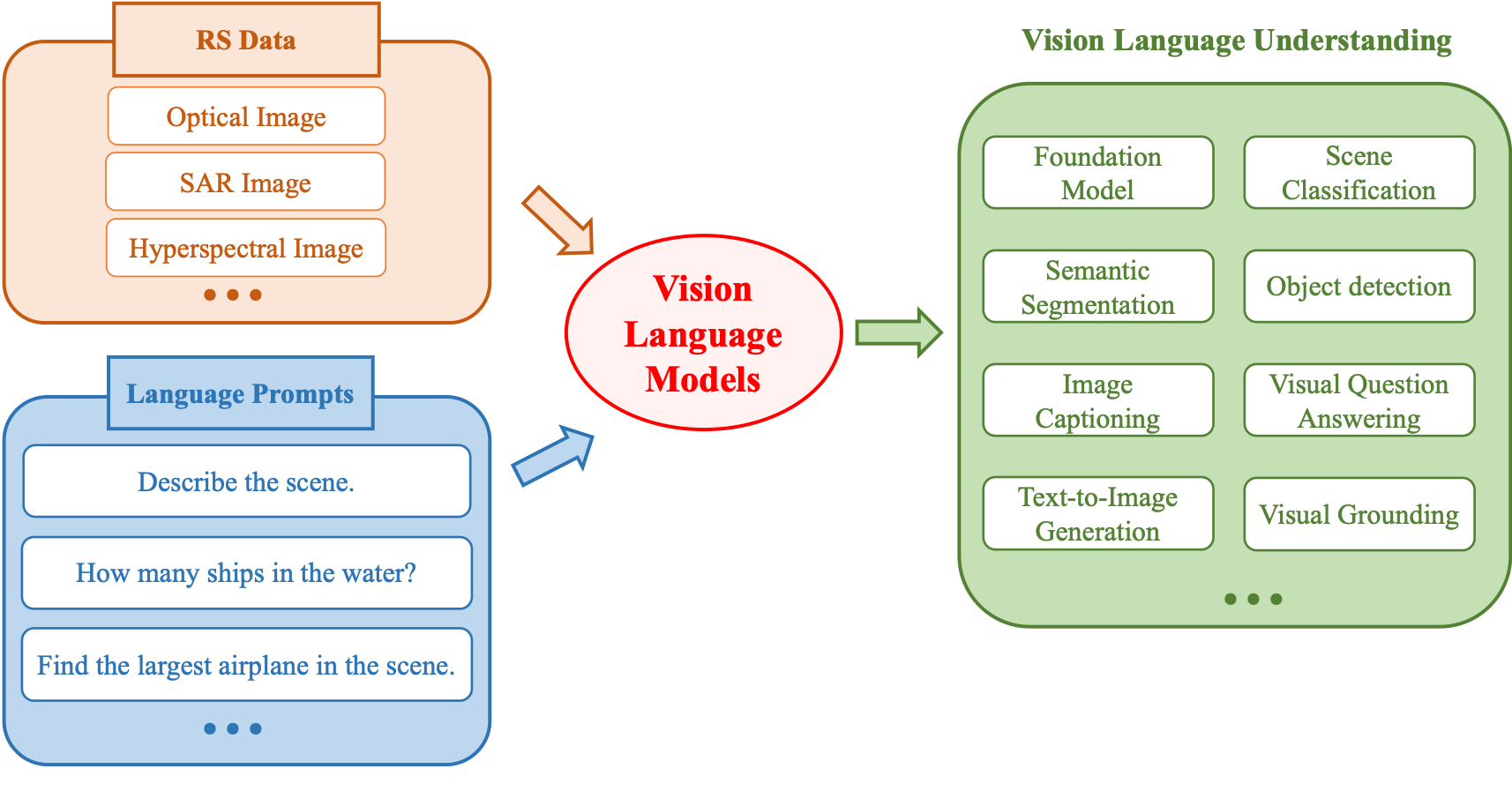}
    \caption{Vision language models in remote sensing.}
    \label{fig:overview}
\end{figure*}

In recent years, Large Language Models (LLMs) have emerged as a popular research topic in the fields of Natural Language Processing (NLP) and computer vision. These models build large-scale transformer networks for natural language understanding and have achieved state-of-the-art performance in various language understanding tasks, such as language modeling, text generation, and question answering~\cite{radford2018improving,chen2023minigptv2}. Notably, the remarkable achievements of ChatGPT~\cite{chatgpt} have sparked a wave of interest and research in the field of Artificial General Intelligence (AGI). Empowered by world knowledge and advanced reasoning capabilities, these models have demonstrated unparalleled advancements in language understanding. The great success of large language models has encouraged numerous research in Vision-Language Models (VLMs), propelling the advent of AI2.0.

VLMs are generally defined as a family of artificial intelligence models that combine computer vision and natural language processing techniques to formulate a comprehensive understanding of both visual and textual information. With the ability to jointly recognize visual and semantic patterns and their relationships, VLMs can go beyond recognizing the objects in an image and can understand the relationships between them, as well as generate natural language descriptions of the image. This makes them better suited for tasks that require both visual and textual understanding, such as image captioning, text-based image retrieval, and visual question answering. More importantly, by combining vision models with LLMs, VLMs offer a more comprehensive and human-like ability to understand visual content. In recent years, VLMs have demonstrated impressive results in a variety of computer vision tasks, including image understanding~\cite{chen2022visualgpt,zhu2023chatgpt}, visual question answering~\cite{li2022blip,li2023blip}, text-to-image generation~\cite{rombach2022high}, semantic segmentation~\cite{chen2022semi,zhang2023language}, object detection~\cite{zhang2023text,lu2023few}, etc. 

In the field of remote sensing, the utilization of VLMs represents a relatively recent and emerging area of research. VLMs possess a unique combination of visual understanding, human-like world knowledge, and robust reasoning capabilities, which collectively enhance remote sensing data analysis in a more intelligent and human-like manner. In general, VLMs offer several advantages over previous methods that solely rely on visual information. Unlike earlier vision-only models that primarily focus on supervised learning and encounter challenges when dealing with out-of-distribution data, VLMs harness semantic reasoning capabilities to establish connections between visual concepts. Consequently, VLMs demonstrate significantly improved performance when confronted with out-of-distribution data. This breakthrough paves the way for zero-shot/open-vocabulary visual understanding tasks, aligning more effectively with real-world applications where visual objects of interest may belong to previously unseen categories or concepts. Furthermore, VLMs present an opportunity to explore the integration of general and expert knowledge into visual analysis tasks concerning remote sensing data. For instance, VLMs possess the awareness that ships are more likely to be located in water rather than on land. Consequently, an object detection model based on VLMs would tend to refrain from detecting ships on the ground, demonstrating the potential for improved remote sensing data analysis. Moreover, VLMs facilitate the execution of novel tasks that demand both visual and language comprehension, such as image captioning, visual question answering, and language-guided editing. Consider the scenario of employing a VLM-based chatbot, which allows individuals, including non-remote sensing experts, to engage in conversational interactions using everyday language to comprehend and analyze remote sensing data. The prospect of such a system is incredibly promising. Recent work~\cite{osco2023potential} explored the potential of Visual ChatGPT~\cite{wu2023visual}, a cutting-edge LLM founded on the GPT architecture, to build a Chatbot tailored for a range of remote sensing image processing tasks, where an advanced ChatGPT model serves as an adept prompt manager, ensuring a thorough understanding of user prompts and seamlessly mapping them to the appropriate tools or algorithms.

With the growing availability of textual metadata associated with RS data, researchers have started exploring the use of vision and language models in this domain \cite{tuia2021}. In recent years, some early attempts try to explore VLMs for various RS data analysis tasks, including RS image captioning~\cite{shi2017can,lu2017exploring,zhang2017natural,zhang2019description,li2020multi,wang2020word,li2020truncation,zhao2021high,zia2022transforming}, text-based RS image generation~\cite{bejiga2019retro,chen2021remote,zhao2021text,xu2022txt2img}, text-based RS image retrieval~\cite{abdullah2020textrs,rahhal2020deep,yuan2022remote,al2022multilanguage,cheng2021deep,yuan2021lightweight,yuan2022exploring,rahhal2023contrasting}, visual question answering~\cite{lobry2020rsvqa,zheng2021mutual,chappuis2022language,al2022open,bazi2022bi,yuan2022easy}, scene classification~\cite{li2017zero,sumbul2017fine,quan2018structural,wang2021distance,li2022generative}, semantic segmentation~\cite{chen2022semi,zhang2023language}, object detection~\cite{jiang2022few,zhang2023text,lu2023few}, etc. Notably, RS5M~\cite{zhang2023rs5m} introduced a large-scale remote sensing image caption dataset by carefully filtering RS-related images from publicly available datasets and leveraged the BLIP-2 model~\cite{li2023blip} for automatic image caption generation. Pioneering work RSGPT~\cite{hu2023rsgpt} built a high-quality human-annotated Remote Sensing Image Captioning dataset (RSICap) that facilitates the development of large VLMs in the RS field, along with a GPT-based model for remote sensing image captioning and visual question answering. With the increasing availability of large-scale RS datasets and advances in deep learning techniques, the use of VLMs is expected to play a significant role in the future of RS applications.

In this work, we present a comprehensive review of the evolution of models in RS from vision to language and to VLMs. Specifically, we conducted an extensive literature survey on the recent advancements in VLMs in RS. Furthermore, we provide valuable insights and recommendations for potential future research directions in the domain of VLMs for RS applications. Our work contributes to a better understanding of the current state-of-the-art in VLMs and provides instructions for researchers in this field to explore the potential of these models in RS tasks.

\section{From Vision-Centric to Vision-Language Models}

\subsection{Vision-Centric Models}
The most commonly used vision-centric models are convolutional neural networks (CNNs). CNNs have become one of the most popular and successful vision models due to their ability to extract high-level features by performing convolution operations on input images, followed by pooling and non-linear activation functions. These models are typically trained using backpropagation, a form of gradient descent, to minimize the error between the predicted output and the ground truth label.

CNNs have a long history, dating back to the 1980s. However, it was not until 1998 that CNNs were first used for image classification tasks. In this work, LeCun et al.~\cite{lenet} proposed the LeNet-5 architecture, which consisted of multiple convolutional layers followed by fully connected layers. The LeNet-5 architecture achieved state-of-the-art results on the MNIST dataset, which consists of handwritten digit recognition.
Since then, several advancements have been made in CNN architectures. For example, the AlexNet architecture~\cite{alexnet}, proposed in 2012, achieved state-of-the-art results on the ImageNet~\cite{deng2009imagenet} dataset, which consists of more than one million images in 1,000 classes. AlexNet used a deeper network with smaller filters and a larger number of hidden units, which allowed it to capture more complex features in images.
In 2014, the VGG architecture~\cite{vggnet} achieved similar performance to AlexNet on the ImageNet dataset while using fewer parameters.
In 2015, the GoogLeNet architecture~\cite{googlenet}, also known as Inception, was proposed. This architecture used a module called Inception, which consisted of multiple convolutional filters with different sizes in parallel, allowing the network to capture features at different scales. 
In 2016, the ResNet architecture~\cite{resnet} was proposed, which introduced residual connections to address the problem of vanishing gradients in deep networks. ResNet used skip connections that allowed the network to learn residual functions instead of directly learning the underlying mapping, which made training deeper networks easier. ResNet achieved state-of-the-art results on the ImageNet dataset with a very deep network of 152 layers.
In 2016, DenseNet architecture~\cite{densenet} was proposed to introduce dense connections between layers, where each layer is connected to all subsequent layers in a feed-forward fashion. This approach enables feature reuse and promotes gradient flow, resulting in improved performance with fewer parameters than traditional deep neural networks.
In 2017, the ResNeXt architecture~\cite{resnext} was proposed to use a cardinality parameter to increase model capacity without significantly increasing computational complexity. ResNeXt demonstrated the effectiveness of using parallel paths with different filter sizes and numbers of channels to improve the capacity and accuracy of deep convolutional neural networks.
In 2019, EfficientNet~\cite{efficientnet} was proposed to use a combination of compound scaling, efficient block structures, and neural architecture search to achieve state-of-the-art performance on image recognition tasks while maintaining a small number of parameters and computational requirements.

More recently, transformer-based models, initially developed for natural language processing tasks, have been widely explored in numerous computer vision tasks. These models, known as vision transformers, use a self-attention mechanism to extract features from images, allowing them to learn global dependencies between different regions of the image.
The self-attention mechanism is formulated as
\begin{equation}
    \text{Attention}(Q,K,V)=\text{Softmax}(\frac{QK^T}{\sqrt{d_k}})V,
\end{equation}
where attention weights are computed by performing a dot-product operation between the query $Q$ and the key $K$, and a scaling factor $\sqrt{d_k}$ and a softmax operation are applied to the normalize the attention weights. The resulting weights are multiplied by the corresponding value feature $V$ to generate the final output features.

\begin{figure}
    \centering
    \includegraphics[width=8cm]{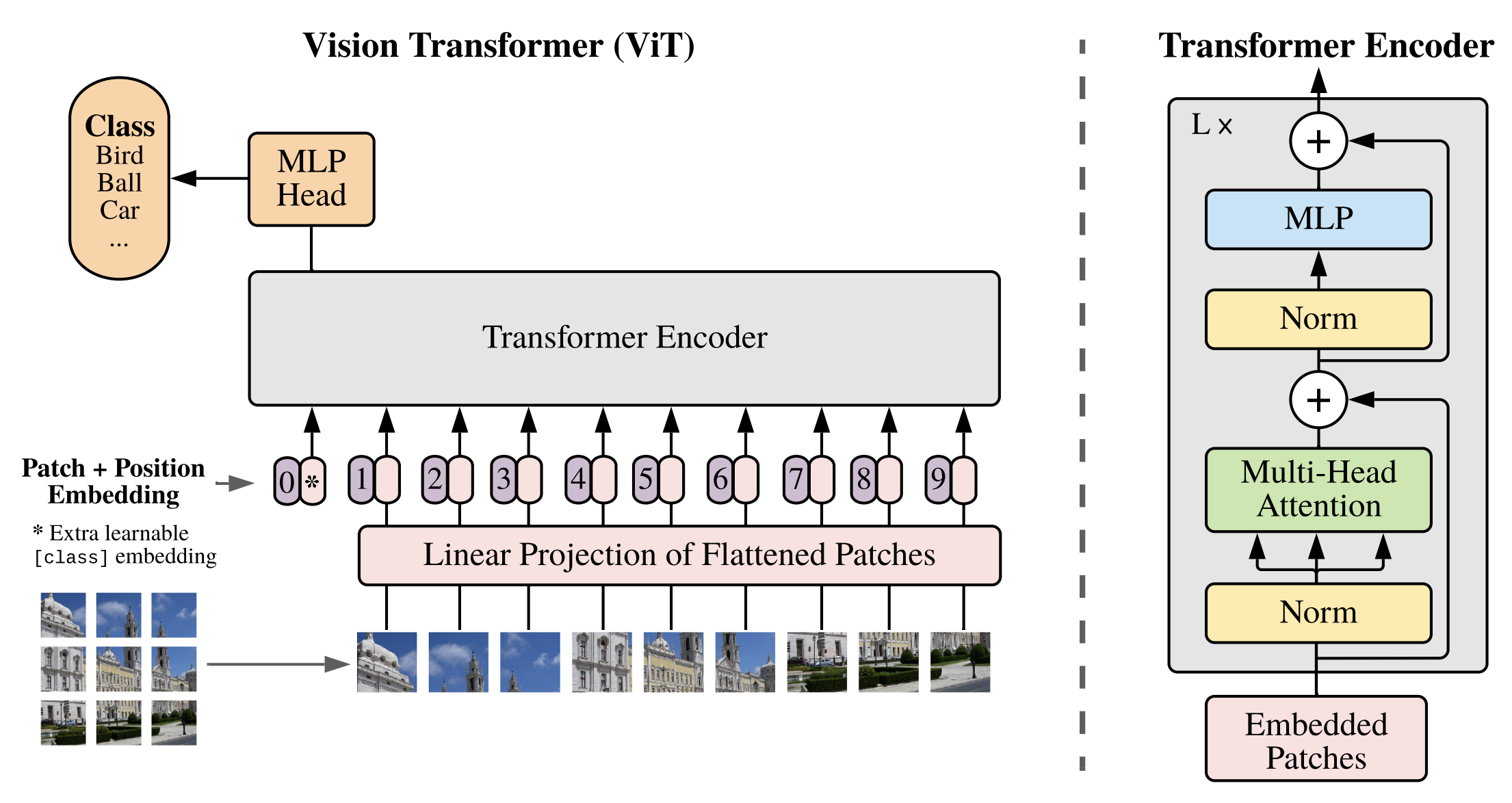}
    \caption{Architecture overview of ViT model~\cite{vit}}
    \label{fig:vit}
\end{figure}

The first transformer-based architecture for image classification is ViT proposed in~\cite{vit}. Fig. \ref{fig:vit} shows the overview of the ViT model. 
Several variants of the ViT architecture have been proposed, including DeiT~\cite{deit}, TNT~\cite{tnt}, and PVT~\cite{pvt}, which further improved performance by incorporating techniques such as distillation, token mixing, and pyramid vision transformer.
In addition, considering that the coarse patchify process in ViT neglects the local image information, the Shifted windows Transformer (Swin)~\cite{swin} was proposed to utilize a shifted window along the spatial dimension to model the global and boundary features.
Twins~\cite{twins} adopted a spatially separable self-attention mechanism, similar to depth-wise convolution~\cite{mobilenetv2}, to model the local-global representation.
ViL~\cite{vil} replaced the single class token with a set of local embeddings, which performed inner attention and interaction with their corresponding 2D spatial neighbors.
VOLO~\cite{volo} introduced outlook attention, similar to a patch-wise dynamic convolution, to focus on the finer-level features through unfold, linear-wights attention, and refold operations.

\subsection{Large Language Models}
Large language models (LLMs) have emerged as a popular research topic in the fields of natural language processing (NLP) and computer vision in recent years. These models build large-scale transformer networks for vision and natural language understanding and have achieved state-of-the-art performance in various language-related tasks, such as language modeling, text generation, and question answering~\cite{radford2018improving,devlin2018bert}. In this section, we will provide an overview of some of the critical developments in large language models.

\begin{figure}[htp]
    \centering
    \includegraphics[width=4cm]{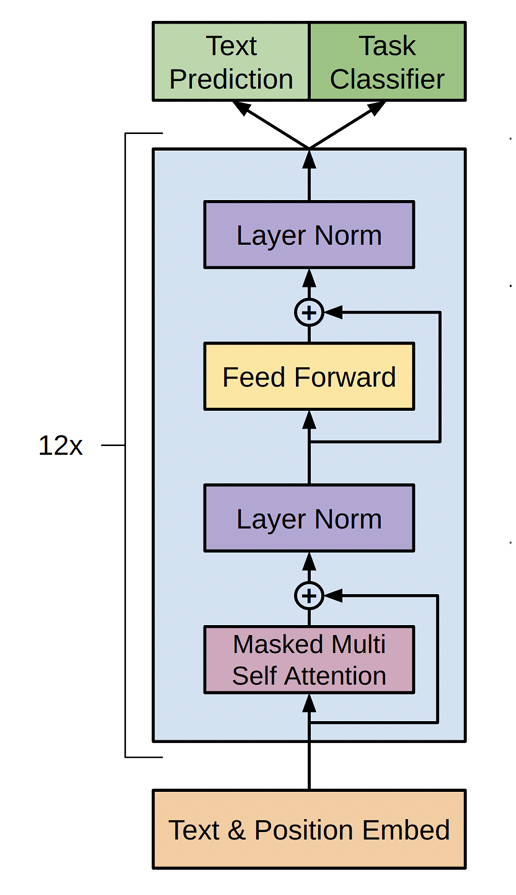}
    \caption{Network architecture and training objectives of the GPT model~\cite{radford2018improving}.}
    \label{fig_gpt}
\end{figure}

One of the pioneering and well-known large language models is the GPT (Generative Pre-trained Transformer)~\cite{radford2018improving} model developed by OpenAI. The GPT model was trained on a massive corpus of web text and achieved impressive performance on a wide range of language modeling and text generation tasks. Fig. \ref{fig_gpt} gives an overview of the GPT method. Given a sequence of tokens $t = \{t_1,...,t_N\}$, GPT maximizes the following likelihood:
\begin{equation}
    \mathcal{L}(t) = \mathbb{E}_{t_i} [\log P(t_i|t_{i-k},...,t_{i-1}; \theta)],
\end{equation}
where $k$ denotes the length of the context window, and $\theta$ denotes the network parameters.

Since the GPT model, the GPT series has undergone several iterations. GPT-2~\cite{radford2019language} with 1.5 billion parameters, achieved remarkable performance on numerous language tasks, including language translation, summarization, question answering, and text completion, and has gained widespread attention for its ability to generate high-quality, coherent, and fluent text that looks indistinguishable from text written by humans. GPT-3~\cite{brown2020language} shows that pretrained large language models can be zero-shot learners and ignited research enthusiasm in in-context learning. InstructGPT~\cite{ouyang2022training}, one of the key techniques used in ChatGPT, introduced a promising approach for improving the control and flexibility of large language models. It works by adding high-level instructions in the form of natural language phrases or templates. These instructions can be used to guide the generation process and ensure that the output text satisfies specific constraints or requirements. The latest version, GPT-4~\cite{openai2023gpt4} demonstrated the enormous potential of large-scale language models for advancing AI and has paved the way for a new era of intelligent machines that can understand and communicate with us in a more natural way.

\begin{figure}
    \centering
    \includegraphics[width=8cm]{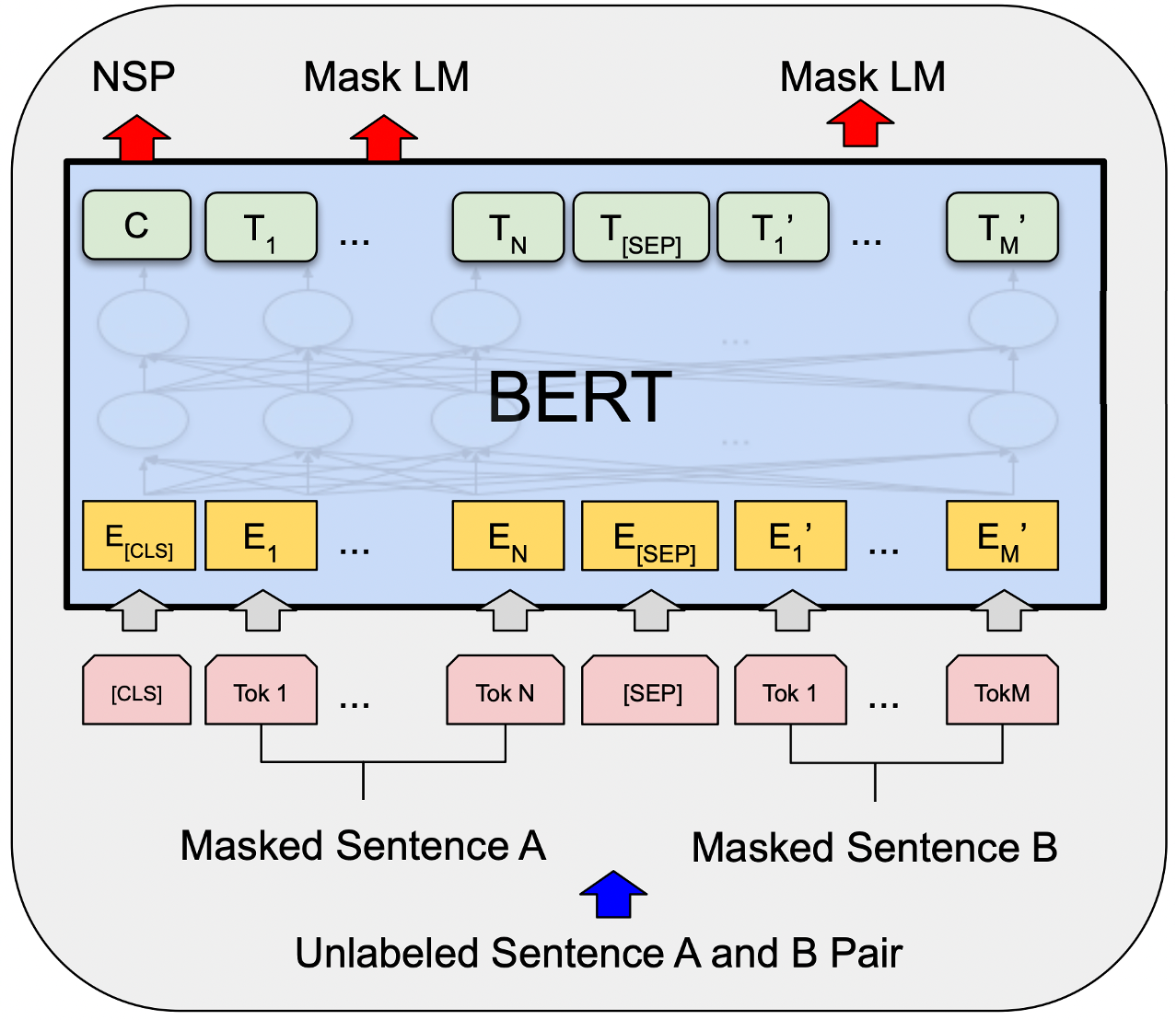}
    \caption{Pre-training objectives of the BERT model~\cite{liu2019roberta}.}
    \label{fig_bert}
\end{figure}

Another popular large language model is BERT~\cite{devlin2018bert}, developed by Google. Unlike the GPT models, BERT was pre-trained on a bidirectional task, meaning that it can take into account both the left and right contexts of a word during pre-training. This enables BERT to capture more nuanced relationships between words and achieve state-of-the-art results on various NLP tasks. Specifically, the pre-training objectives of BERT contain two unsupervised tasks: 1) masked LM, which tries to predict several masked tokens given nearby tokens; 2) Next Sentence Prediction, which tries to predict the next sentence given the previous sentence, as shown in Fig. \ref{fig_bert}. Given a sequence of tokens $t = \{t_1,...,t_N\}$, the training objective of masked LM is defined as: 
\begin{equation}
    \mathcal{L}(t) = \mathbb{E}_\mathcal{M} [\sum_{i \in \mathcal{M}} \log P(t_i|t^{\mathcal{M}}; \theta)], \\
\end{equation}
where $\mathcal{M}$ denotes randomly masked positions, $t^{\mathcal{M}} = \{t_i : i \notin \mathcal{M}\}_{i=1}^N$ represents the corrupted sequence that is masked according to $\mathcal{M}$, and $\theta$ denotes the network parameters.

There are numerous variants of the BERT model. For example, RoBERTa~\cite{liu2019roberta}, developed by Facebook AI, improved BERT by using more iterations and data augmentation techniques, including dynamic masking and noising. ALBERT~\cite{lan2019albert} proposed three techniques, including factorized embedding parameterization, cross-layer parameter sharing, and sentence order prediction task, to reduce the model size of BERT and improve model training speed. MacBERT~\cite{cui2020revisiting} proposed to replace the [MASK] token with another synonym at mask locations instead of using the [MASK] tag. 

In addition to GPT and BERT, several other large language models have emerged in recent years. T5~\cite{raffel2020exploring} applied a single unified architecture to a wide range of NLP tasks by task-agnostic pre-training on a massive corpus of diverse text data with the goal of creating a general-purpose language model. CoT~\cite{wei2022chain} proposed a useful technique called chain-of-thought prompting that enables intermediate reasoning in large language models.

\subsection{Vision-Language Models}

Given the success of pre-trained models in computer vision and natural language processing (NLP), researchers have attempted to pre-train large-scale models that incorporate both modalities, which are called Vision-Language Models (VLMs). These VLMs can be categorized into two model architectures: fusion-encoder and dual-encoder models. Fusion-encoder models use a multi-layer cross-modal Transformer encoder to jointly encode image and text pairs and fuse their visual and textual representations. Meanwhile, dual-encoder models encode images and text separately and use either a dot product or MLP to capture the interactions between the modalities.

\begin{figure}
    \centering  \includegraphics[width=9cm]{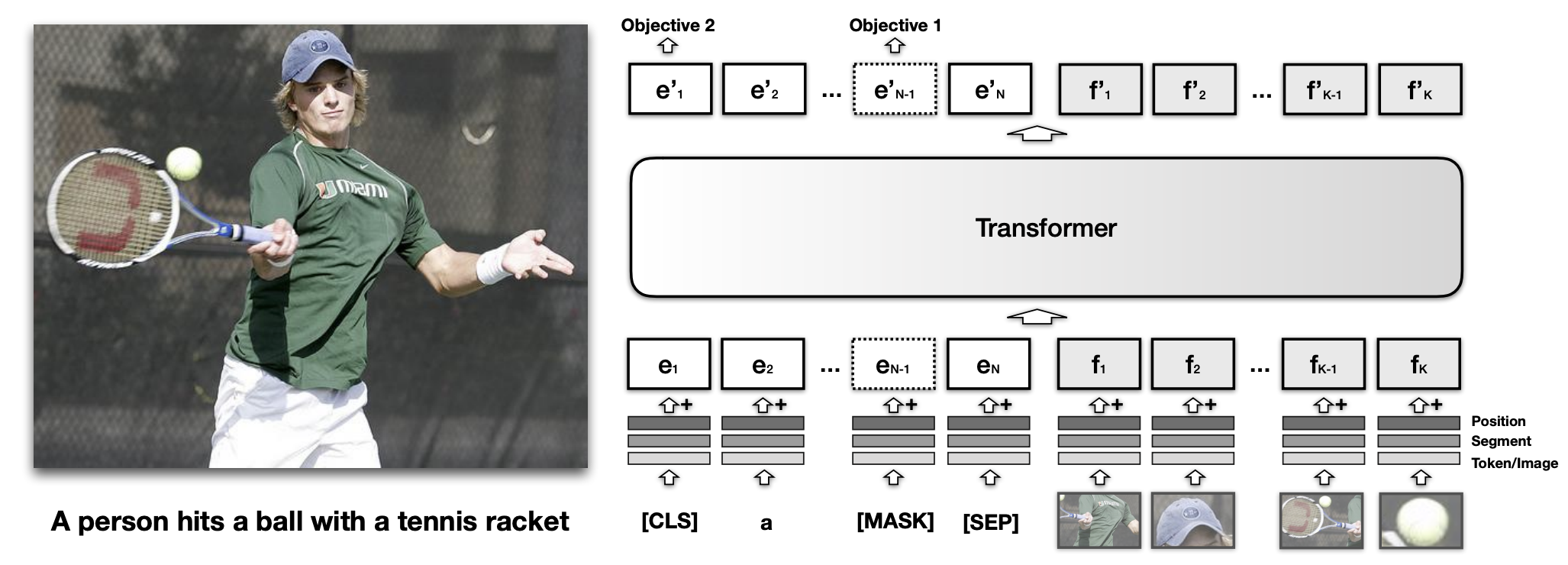}
    \caption{Illustration of the architecture of VisualBERT~\cite{li2019visualbert}}
    \label{fig:visualBERT}
\end{figure}

\subsubsection{Fusion Encoder}
The fusion encoder accepts visual features and text embeddings as input and employs multiple fusion techniques to capture the interaction between visual and text modalities. The latent features of the final layer are regarded as the fused representation of the distinct modalities after either a self-attention or cross-attention operation. VisualBERT~\cite{li2019visualbert} is a pioneering work that implicitly aligned elements of an input text with regions in an associated input image using self-attention. It combined BERT~\cite{devlin2018bert} for processing natural language and pretrained Faster-RCNN~\cite{faster-rcnn} for generating object proposals. The original text, along with the image features extracted from object proposals, were treated as unordered input tokens and fed into VisualBERT to capture the intricate associations by jointly processing them with multiple Transformer layers (See Fig.~\ref{fig:visualBERT}). Subsequently, several VLM models, including Uniter~\cite{chen2020uniter}, OSCAR~\cite{li2020oscar}, InterBert~\cite{lin2020interbert}, utilized BERT as a text encoder and Faster-RCNN as an object proposal generator to model vision-language interaction.

\begin{figure*}
    \centering  \includegraphics[width=17cm]{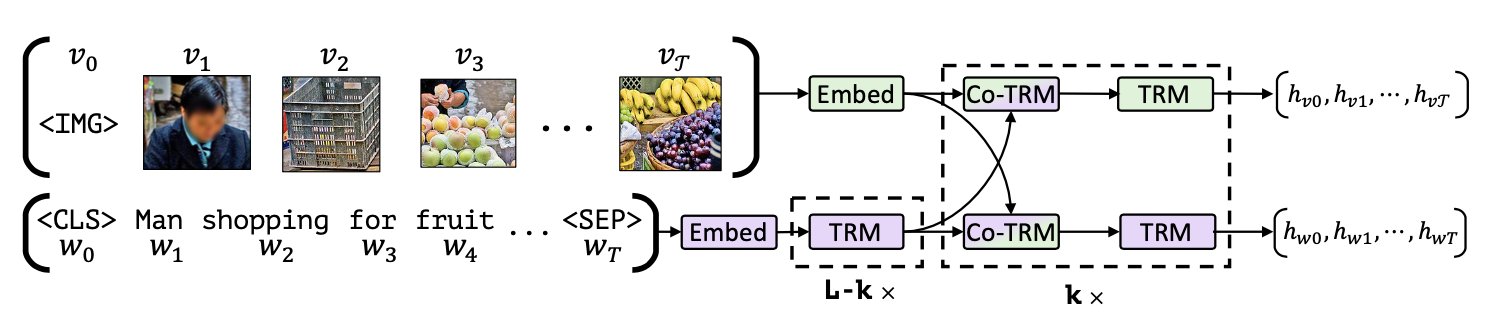}
    \caption{Illustration of the architecture of ViLBERT~\cite{lu2019vilbert}.}
    \label{fig:vilBERT}
\end{figure*}

In contrast to self-attention operations utilized in single-stream architectures, dual-stream architectures utilize a cross-attention mechanism to capture the interaction between visual and language modalities. The cross-attention layer usually consists of two unidirectional sub-layers, with one processing language-to-vision and the other processing vision-to-language. These sub-layers facilitate information exchange and semantic alignment between the two modalities. One prominent example is ViLBERT~\cite{lu2019vilbert}, which processed visual and textual inputs separately and employed co-attentional transformer layers to enable information exchange between modalities. Fig.~\ref{fig:vilBERT} illustrates how each stream is composed of transformer layers (TRM) and novel co-attention transformer blocks (Co-TRM). In addition, recent works such as  LXMERT~\cite{tan2019lxmert}, Visual Parsing~\cite{xue2021probing}, ALBEF~\cite{li2021align} and WenLan~\cite{huo2021wenlan} also employed separate transformers before cross-attention to decouple intra-modal and cross-modal interaction. Chen et al.~\cite{chen2022visualgpt} proposed VisualGPT that adapts pre-trained language models to small quantities of in-domain image-text data by utilizing a novel self-resurrecting encoder-decoder attention mechanism.

\begin{figure*}
    \centering
    \includegraphics[width=16cm]{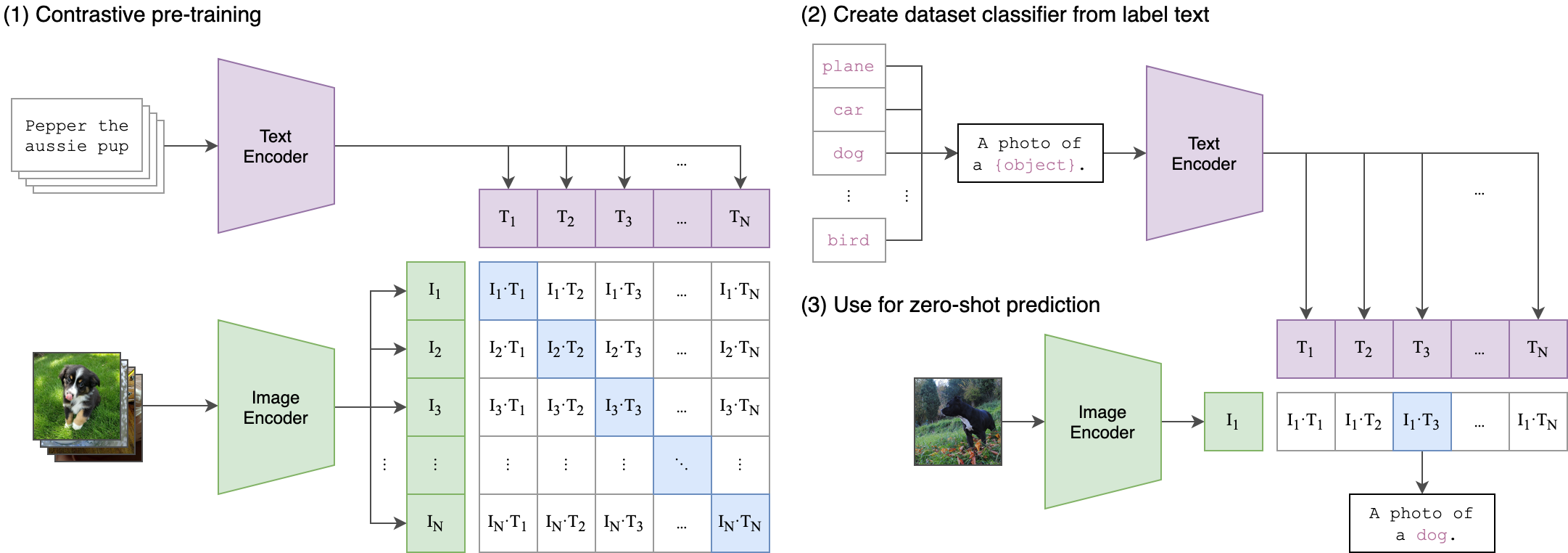}
    \caption{Illustration of the architecture of CLIP~\cite{radford2021learning}.}
    \label{fig:CLIP}
\end{figure*}

\subsubsection{Dual Encoder}
A dual encoder uses two separate encoders for each modality to encode visual and textual information independently. The image and text embeddings encoded from the corresponding encoder are then projected onto the shared semantic latent space via operations such as attention layers or dot products, which are used to calculate similarity scores between vision and language embeddings. Compared to the fusion encoder mechanism, the dual encoder mechanism does not use the complex cross-attention in the transformer, because it pre-computes and stores the image and text embeddings, making vision-language interaction modeling more efficient. For example, Contrastive Language-Image Pre-training (CLIP)~\cite{radford2021learning},  as illustrated in Fig.~\ref{fig:CLIP}, utilizes a text encoder and an image encoder jointly to accurately match pairings of (image, text) samples. Let ${{(x_i^I, x_i^T)}}$ be a batch of $N$ image-text pairs, and $z_i^I$ and $z_j^T$ denote the normalized embeddings of the $i_{th}$ image and $j_{th}$ text, respectively. CLIP employs InfoNCE~\cite{oord2018representation} to calculate the loss, as shown in Eq.~\ref{eq:infonce}.

\begin{equation}
\label{eq:infonce}
\begin{aligned}
    L_{I2T} = - \frac{1}{N} \sum_{i=1}^{N} \log \frac{\exp({sim}( z_i^I, z_i^T)/\tau)}{\sum_{j=1}^{N}\exp({sim}(z_i^I, z_j^T)/\tau)}, \\
    L_{T2I} = - \frac{1}{N} \sum_{i=1}^{N} \log \frac{\exp({sim}( z_i^T, z_i^I)/\tau)}{\sum_{j=1}^{N}\exp({sim}(z_i^T, z_j^I)/\tau)},
\end{aligned}
\end{equation}
where $\mathrm{sim}(,)$ is the similarity function calculated using dot product, and $\tau$ is a temperature variable used to scale the logits. The loss function is symmetrical for both the image and text encoder, and the final loss is the average of image-to-text matching loss and text-to-image matching loss.

Furthermore, ALIGN~\cite{jia2021scaling} introduced a dual-encoder architecture with a contrastive loss to align image and text representations before cross-modal attention. This loss pulls together matched image-text pairs and pushes apart non-matched pairs. To overcome the limitations of CLIP, Li et al.~\cite{li2021supervision} proposed a novel training paradigm called data-efficient CLIP, which improves the efficiency of learning generic visual features. This approach incorporated both intra-modality self-supervision as well as inter-modality multi-view supervision and further introduced a supervision signal based on finding similar texts via their embeddings. Alayrac et al. introduced Flamingo~\cite{alayrac2022flamingo}, a model trained on only a few input/output examples that can accept interleaved visual data and text as input and generate text in an open-ended manner. Specifically, Flamingo used spatiotemporal features from the vision encoder and initialized cross-attention layers to interleave these visual tokens between the pretrained language model layers. Similarly, Li et al. proposed a generic and efficient VLM model, named BLIP-2~\cite{li2023blip}, which bootstrapped vision-language pretraining from off-the-shelf frozen pre-trained image encoders and frozen large language models. They first bootstrapped vision-language representation learning from a frozen image encoder and then bootstrapped vision-to-language generative learning from a frozen language model. Recently, Kirillov et al. built a foundation segmentation model (SAM)~\cite{kirillov2023segment} that can be applied to a range of image understanding tasks, such as semantic segmentation, edge detection, panoptic segmentation, and instance segmentation, in a zero-shot setting. The SAM adopted a heavyweight image encoder to obtain image embeddings and then used a prompt encoder to generate queries from input prompts to produce multiple object masks and confidence scores. Zhu et al. proposed MiniGPT-4~\cite{zhu2023minigpt} that trained a single linear projection layer to align the visual features encoded from a pretrained ViT with a pretrained large language model. This simplified model with only 13B parameters shows impressive performance on various cross-modality tasks, such as image captioning, writing poems/stories from images, and building websites from draft images. Additionally, InstructBLIP~\cite{dai2023instructblip}, LLAVA~\cite{liu2024visual}, and VisionLLM~\cite{wang2023visionllm} introduced general-purpose vision-language models based on instruction tuning. Kosmos~\cite{huang2023language,peng2023kosmos} introduced a Multimodal Large Language Model (MLLM) trained from scratch on web-scale multimodal corpora. This model was designed to perceive various modalities, learn in context (i.e., in a few-shot manner), and follow instructions (i.e., in a zero-shot fashion).

\begin{table*}[!ht]
    \centering
    \resizebox{\textwidth}{0.45\textheight}{%
    \begin{tabular}{p{2.5cm}p{1cm}p{1cm}p{3cm}p{3cm}p{5.5cm}}
        \hline
        Method & Year & Task & Visual Encoder & Text Encoder & Datasets \\
        \hline
        \rowcolor{LightYellow}
        RingMo~\cite{ringmo} & 2022 & FM & ViT~\cite{vit}, Swin Transformer~\cite{swin} & - & two million RS images collected from multiple public datasets and GF-2  \\
         \rowcolor{LightYellow}
        Wang et al.~\cite{advancing} & 2022 & FM & ViT~\cite{vit}, ViTAE~\cite{vitae} & - & MillionAID~\cite{million-aid} \\
         \rowcolor{LightYellow}
        GFM~\cite{gfm} & 2023 & FM & Swin-B~\cite{swin} & - & GeoPile~\cite{gfm} \\
        \rowcolor{LightYellow}
        Cha et al. & 2023 & FM & ViT \cite{vit} & - & MillionAID \cite{million-aid} \\
        \rowcolor{LightRed}
        Shi et al.~\cite{shi2017can} & 2017 & RSIC & VGG-f~\cite{chatfield2014return} & - & Google Earth, GF-2 \\
        \rowcolor{LightRed}
        Lu et al.~\cite{lu2017exploring} & 2017 & RSIC & AlexNet~\cite{alexnet}, VGG~\cite{vggnet}, GoogLeNet~\cite{googlenet} & LSTM~\cite{hochreiter1997long} & RSICD~\cite{lu2017exploring}\\
        \rowcolor{LightRed}
        Zhang et al.~\cite{zhang2017natural} & 2017 & RSIC & CaffeNet~\cite{alexnet} & RNN~\cite{hopfield1985neural} & UCM~\cite{yang2010bag}\\
        \rowcolor{LightRed}
        VAA~\cite{zhang2019vaa} & 2019 & RSIC & VGG16~\cite{vggnet} & LSTM~\cite{hochreiter1997long} & UCM-Captions~\cite{qu2016deep}, Sydney-Captions~\cite{qu2016deep}\\
        \rowcolor{LightRed}
        Zhang et al.~\cite{zhang2019description} & 2019 & RSIC & VGG16~\cite{vggnet} & LSTM~\cite{hochreiter1997long} & UCM-Captions~\cite{qu2016deep},Sydney-Captions~\cite{qu2016deep}, RSICD~\cite{lu2017exploring}\\
        \rowcolor{LightRed}
        Li et al.~\cite{li2020multi} & 2020 & RSIC & ResNet-101~\cite{resnet} & LSTM~\cite{hochreiter1997long} & UCM-Captions~\cite{qu2016deep},Sydney-Captions~\cite{qu2016deep}, RSICD~\cite{lu2017exploring}\\
        \rowcolor{LightRed}
        VRTMM~\cite{shen2020remote} & 2020 & RSIC & VGG16~\cite{vggnet} & Transformer~\cite{vaswani2017attention} & NWPU-RESISC45~\cite{cheng2017remote}\\
        \rowcolor{LightRed}
        Wang et al.~\cite{wang2020word} & 2020 & RSIC & AlexNet~\cite{alexnet}, VGG~\cite{vggnet}, ResNet~\cite{resnet} & Transformer~\cite{vaswani2017attention} & UCM-Captions~\cite{qu2016deep}, Sydney-Captions~\cite{qu2016deep}, RSICD~\cite{lu2017exploring}\\
        \rowcolor{LightRed}
        Li et al.~\cite{li2020truncation} & 2020 & RSIC & AlexNet~\cite{alexnet}, VGG~\cite{vggnet}, ResNet~\cite{resnet},  GoogleNet~\cite{googlenet} & LSTM~\cite{hochreiter1997long} & UCM-Captions~\cite{qu2016deep},Sydney-Captions~\cite{qu2016deep}, RSICD~\cite{lu2017exploring}\\
        \rowcolor{LightRed}
        Zhao et al.~\cite{zhao2021high} & 2021 & RSIC & ResNet-50~\cite{resnet} & LSTM~\cite{hochreiter1997long} & UCM-Captions~\cite{qu2016deep},Sydney-Captions~\cite{qu2016deep}, RSICD~\cite{lu2017exploring}\\
        \rowcolor{LightRed}
        Zia et al.~\cite{zia2022transforming} & 2022 & RSIC & ResNet~\cite{resnet} & Transformer~\cite{vaswani2017attention} & UCM-Captions~\cite{qu2016deep},Sydney-Captions~\cite{qu2016deep}, RSICD~\cite{lu2017exploring}\\
        \rowcolor{LightRed}
        VLCA.~\cite{wei2023vlca} & 2023 & RSIC & CLIP~\cite{radford2021learning} & GPT-2~\cite{radford2019language} & DIOR-Captions~\cite{wei2023vlca} \\
        \rowcolor{LightOrange}
        Bejiga et al.~\cite{bejiga2019retro} & 2019 & T2IG & GAN~\cite{creswell2018generative} & - & MODIS  \\
        \rowcolor{LightOrange}
        BTD-sGAN~\cite{chen2021remote} & 2021 & T2IG & D-sGAN~\cite{lv2021remote} & - &  Oxford-102, GF  \\
        \rowcolor{LightOrange}
        StrucGAN~\cite{zhao2021text} & 2021 & T2IG & AttnGAN~\cite{xu2018attngan} & LSTM~\cite{hochreiter1997long} & RSICD~\cite{lu2017exploring} \\
        \rowcolor{LightOrange}
        Txt2Img-MHN~\cite{xu2022txt2img} & 2022 & T2IG & VQVAE~\cite{van2017neural},  VQGAN~\cite{esser2021taming} & BPE~\cite{sennrich2015neural} & RSICD~\cite{lu2017exploring} \\
        \rowcolor{LightGreen}
        DBTN~\cite{abdullah2020textrs} & 2020 & T2IR & VGG~\cite{vggnet}, Inception V3~\cite{szegedy2016rethinking}, ResNet-50~\cite{resnet}, EfficientNet~\cite{efficientnet} & LSTM~\cite{hochreiter1997long} & TextRS~\cite{abdullah2020textrs} \\
        \rowcolor{LightGreen}
        Rahhal et al.~\cite{rahhal2020deep} & 2020 & T2IR & BiT~\cite{kolesnikov2020big} & Bi-LSTM~\cite{graves2005framewise}   & TextRS~\cite{abdullah2020textrs}, UCM~\cite{yang2010bag} \\
        \rowcolor{LightGreen}
        LW-MCR~\cite{yuan2021lightweight} & 2021 & T2IR & SqueezeNet~\cite{iandola2016squeezenet} & Group CNN& Sydney~\cite{zhang2014saliency},UCM~\cite{yang2010bag}, RSITMD~\cite{yuan2022exploring},  RSICD~\cite{lu2017exploring}\\
        \rowcolor{LightGreen}
        SAM~\cite{cheng2021deep} & 2021 & T2IR & Inception V3~\cite{szegedy2016rethinking} & Bi-GRU~\cite{schuster1997bidirectional} & UCM-Captions~\cite{qu2016deep},Sydney-Captions~\cite{qu2016deep}, RSICD~\cite{lu2017exploring}, NWPU-RESISC45-Captions\\ 
        \rowcolor{LightGreen}
        GaLR~\cite{yuan2022remote} & 2022 & T2IR &  CNN~\cite{yuan2022remote} & GRU~\cite{cho2014learning} & RSICD~\cite{lu2017exploring}, RSITMD~\cite{yuan2022exploring}\\
        \rowcolor{LightGreen}
        Rahhal et al.~\cite{al2022multilanguage} & 2022 & T2IR &  ViT~\cite{vit} & Transformer~\cite{vaswani2017attention} & RSICD~\cite{lu2017exploring}, RSITMD~\cite{yuan2022exploring}, UCM~\cite{yang2010bag} \\
        \rowcolor{LightGreen}
        AMFMN~\cite{yuan2022exploring} & 2022 & T2IR & ResNet-18~\cite{resnet} & GRU~\cite{cho2014learning} & Sydney~\cite{zhang2014saliency},UCM~\cite{yang2010bag}, RSITMD~\cite{yuan2022exploring}, RSICD~\cite{lu2017exploring} \\
        \rowcolor{LightGreen}
        Rahhal et al.~\cite{rahhal2023contrasting} & 2023 & T2IR &  Transformer~\cite{vaswani2017attention} &  Transformer~\cite{vaswani2017attention} & TextRS~\cite{abdullah2020textrs}, UCM~\cite{yang2010bag},Sydney~\cite{zhang2014saliency}, RSICD~\cite{lu2017exploring}\\
        \rowcolor{LightRed}
        Lobry et al.~\cite{lobry2020rsvqa} & 2020 & VQA & ResNet-152~\cite{resnet} & LSTM~\cite{hochreiter1997long} & RSVQA~\cite{lobry2020rsvqa} \\
        \rowcolor{LightRed}
        Zheng et al.~\cite{zheng2021mutual} & 2021 & VQA & VGG-16~\cite{vggnet} & GRU~\cite{cho2014learning} & RSIVQA~\cite{zheng2021mutual} \\
        \rowcolor{LightRed}
        Chappuis et al.~\cite{chappuis2022language} & 2022 & VQA & ResNet-152~\cite{resnet} & BERT~\cite{devlin2018bert} & RSVQA~\cite{lobry2020rsvqa} \\
        \rowcolor{LightRed}
        Al et al.~\cite{al2022open} & 2022 & VQA & Vision Transformer & Transformer~\cite{vaswani2017attention} & VQA-TextRS~\cite{al2022open} \\
        \rowcolor{LightRed}
        Bazi et al.~\cite{bazi2022bi} & 2022 & VQA & ViT~\cite{vit} & CLIP~\cite{radford2021learning} & Sentinel-2 and aerial images \\
        \rowcolor{LightRed}
        Yuan et al.~\cite{yuan2022easy} & 2022 & VQA & CNN~\cite{yuan2022easy} &  RNN~\cite{hopfield1985neural} & RSVQA~\cite{lobry2020rsvqa}, RSIVQA~\cite{zheng2021mutual} \\ 
        \rowcolor{LightRed}
        Yuan et al.~\cite{yuan2022change} & 2022 & VQA & ViT~\cite{vit} & RNN~\cite{hopfield1985neural} & CDVQA~\cite{yuan2022change}\\
        \rowcolor{LightYellow}
        Sun et al.~\cite{sun2022visual} & 2022 & VG & Darknetz\cite{farhadi2018yolov3} & BERT~\cite{devlin2018bert} & RSVG~\cite{sun2022visual} \\
        \rowcolor{LightYellow}
        Zhan et al.~\cite{zhan2023rsvg} & 2023 & VG & CNN & BERT~\cite{devlin2018bert} & DIOR-RSVG~\cite{zhan2023rsvg} \\
        \rowcolor{LightOrange}
        Li et al.~\cite{li2017zero} & 2017 & SC & GoogLeNet~\cite{googlenet} & Word2vec~\cite{mikolov2013efficient} & UCM~\cite{yang2011spatial}, RSSCN7~\cite{zou2015deep}, Sydney 900~\cite{li2017zero}. \\
        \rowcolor{LightOrange}
        Sumbul et al.~\cite{sumbul2017fine} & 2017 & SC & CNN~\cite{sumbul2017fine} & Word2vec~\cite{mikolov2013efficient} &  self collected \\
        \rowcolor{LightOrange}
        Quan et al.~\cite{quan2018structural} & 2018 & SC & GoogLeNet~\cite{googlenet} & Word2vec~\cite{mikolov2013efficient} & UCM~\cite{yang2011spatial}, AID~\cite{xia2017aid}.\\
        \rowcolor{LightOrange}
        Wang et al.~\cite{wang2021distance} & 2021 & SC & ResNet-152~\cite{resnet} & Word2vec~\cite{mikolov2013efficient}  & UCM~\cite{yang2011spatial},  AID~\cite{xia2017aid}, NWPU-45~\cite{cheng2017remote} \\
        \rowcolor{LightOrange}
        Li et al.~\cite{li2022generative} & 2022 & SC & ResNet-101~\cite{resnet} & Word2vec~\cite{wang2016relational}, Fasttest~\cite{joulin2016bag,bojanowski2017enriching}, Glove~\cite{xian2016latent},  BERT~\cite{devlin2018bert} & UCM~\cite{yang2011spatial}, AID-30~\cite{xia2017aid}, NWPU-45~\cite{cheng2017remote} \\
        \rowcolor{LightGreen}
        Zhang et al.~\cite{zhang2023text} & 2023 & OD & Faster RCNN~\cite{faster-rcnn} & Glove~\cite{xian2016latent} & NWPU VHR-10~\cite{cheng2014multi}, DIOR~\cite{li2020object} \\
        \rowcolor{LightGreen}
        Lu et al.~\cite{lu2023few} & 2023 & OD & Faster RCNN~\cite{faster-rcnn} & GRU~\cite{cho2014learning} & NWPU VHR-10~\cite{cheng2014multi},  DIOR~\cite{li2020object}, FAIR1M~\cite{sun2022fair1m} \\
        \rowcolor{LightYellow}
        Chen et al.~\cite{chen2022semi} & 2019 & SS & ResNet~\cite{chen2022semi} & -&  Vaihingen~\cite{rottensteiner2012isprs}, the Zurich Summer dataset\footnotemark[1] \\
        \rowcolor{LightYellow}
        Jiang et al.~\cite{jiang2022few} & 2021 & SS & ResNet~\cite{resnet} & - & ISPRS Vaihingen~\cite{rottensteiner2012isprs}  \\
        \rowcolor{LightYellow}
        Zheng et al. \cite{zhang2023language} & 2023 & SS & 3D-CNN~\cite{zhang2023language} & GPT-2 \cite{radford2019language} & Houston~\cite{debes2014hyperspectral,le20182018}, Pavia~\cite{gamba2004collection}, GID~\cite{tong2020land} \\
        \hline
    \end{tabular}
    }
    \caption{Summary of well-known vision-language models in remote sensing. ``-'' means not applicable. ``FM'' denotes foundation model, ``RSIC" denotes remote sensing image captioning, ``T2IG'' denotes text-based image generation, ``VQA'' denotes visual question answering, ``VG'' denotes visual grounding, ``SC'' denotes scene classification, ``OD'' denotes object detection, and ``SS'' denotes semantic segmentation.}
    \label{tab_summary}
\end{table*}
~\footnotetext[1]{https://sites.google.com/site/michelevolpiresearch/data/zurich-dataset}

\section{Vision-Language Models in Remote Sensing}

\subsection{Vision-Centric Foundation Models}

\begin{figure*}
    \centering
    \includegraphics[width=16cm]{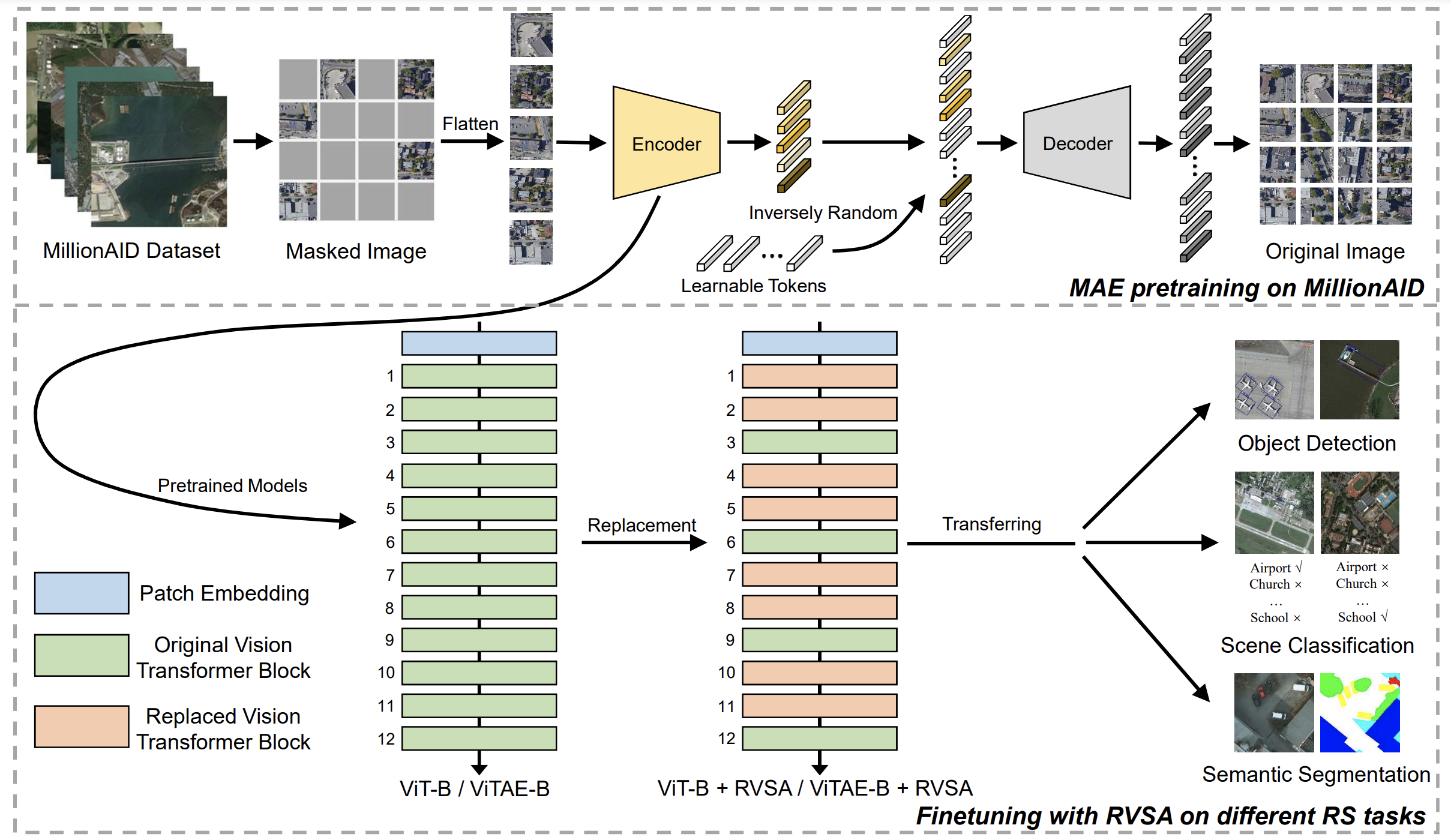}
    \caption{Method overview of the remote sensing vision transformer network in~\cite{advancing}.}
    \label{fig_foundation}
\end{figure*}

Foundation models refer to large pre-trained deep learning neural networks trained in a task-agnostic manner on massive amounts of data. These models can be applied to various downstream tasks using fine-tuning, few-shot, or zero-shot learning. Examples of foundation models include GPT-3~\cite{brown2020language}, BERT~\cite{devlin2018bert}, and T5~\cite{raffel2020exploring}. These models have been pre-trained on large amounts of text data and are able to be finetuned for a wide range of NLP tasks, such as language translation, question answering, and text classification. In remote sensing (RS), pretraining is critical for enhancing the performance on classification, detection, and segmentation tasks~\cite{xiong2022earthnets}. Previous approaches have predominantly leveraged the ImageNet dataset for pertaining. However, transferring the ImageNet pretrained model to RS tasks suffers from huge domain gaps due to the significant difference between natural images and RS images. Thus designing a foundation model tailored for RS data is necessary.

Researchers have pursued this goal using two approaches: supervised learning and self-supervised learning. 
In supervised learning,~\cite{wang2022empirical} pre-trained deep neural networks on the MillionAID dataset, a large-scale RS dataset, and improved the performance on other downstream RS tasks. However, the need for a significant amount of labeled data remains a hurdle, as it can impede the training of larger models.
Therefore, self-supervised techniques gradually become the primary methods used to develop foundation models for remote sensing, as they can leverage a substantial amount of unlabeled data~\cite{wangyi2022ssl}. Early works~\cite{akiva2022self,li2021geographical,ayush2021geography,manas2021seasonal,heidler2023self} resorted to contrastive learning for the foundation model training, incorporating RS-specific information, such as geographic data, time-series data, audio data, and so on. Recently, masked image modeling (MIM) has recently gained increasing attention in computer vision, such as BEiT~\cite{beit}, MAE~\cite{mae}, SimMIM~\cite{simmim}, as it eliminated the need for labels, data augmentation, and selection of positive and negative pairs. Thus it is easier to leverage vast amounts of data. Some works applied MIM to develop RS foundation models. For example,~\cite{ringmo} collected two million RS images from satellite and aerial platforms to create a large-scale RS dataset. Based on the dataset, they designed the first generative self-supervised RS foundation model, RingMo. RingMo achieved state-of-the-art on eight datasets across four downstream tasks, including change detection, scene recognition, object detection, and semantic segmentation.
\cite{advancing} made the first attempt to build a plain vision transformer with about 100 million parameters for a large vision foundation model tailored to RS tasks. The method overview is shown in Fig. \ref{fig_foundation}. They also introduced a rotated varied-size window attention mechanism to enhance the ability of the vision transformer to adapt to RS images. \cite{gfm} discovered that models pretrained on diverse datasets, such as ImageNet-22k, should not be disregarded when constructing geospatial foundation models, as their representations remain effective. Consequently, they built a geospatial foundation model for geospatial applications in a sustainable manner. \cite{billion} developed the first billion-scale foundation model in the RS field and proved the effect of increasing the size of the model from million-scale to billion-scale. A comprehensive literature review of remote sensing foundation models can be found at ~\cite{jiao2023brain}.

 \begin{figure*}[ht]
     \centering
     \includegraphics[width=18cm]{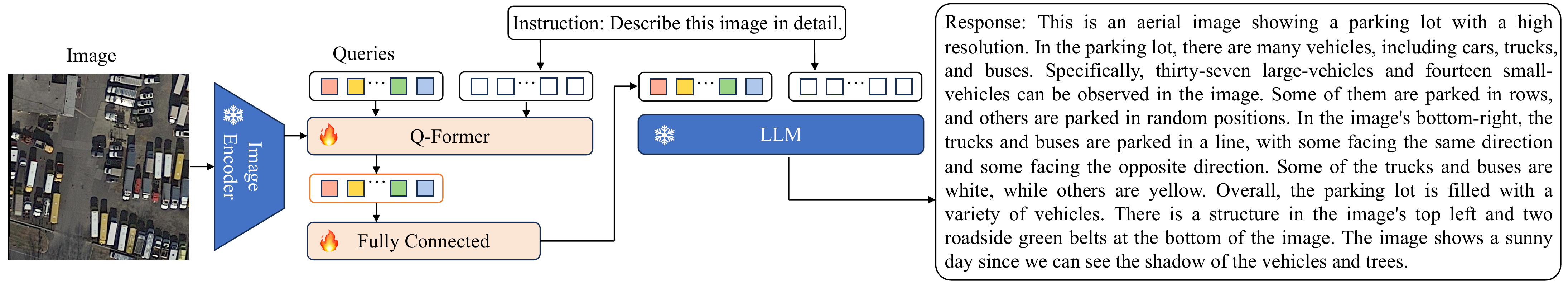}
     \caption{Overview of a remote sensing image captioning method proposed in~\cite{hu2023rsgpt}. The model consists of an image encoder for visual feature extraction, a Q-Former~\cite{dai2023instructblip} for feature compression, a linear projection layer, and a frozen LLM.}
     \label{fig_rsi_captioning}
 \end{figure*}
 
\subsection{Image Captioning}
Remote sensing image captioning (RSIC) is a complex task that requires the machine to understand the content of a remote sensing (RS) image and describe it in natural language. This is a challenging task as the generated description must capture not only the ground elements of different scales but also their attributes and manners in which they interrelate. Unlike other tasks that aim to predict individual tags or words, RSIC aims to generate comprehensive sentences. To generate concise and meaningful sentence descriptions, it is important to identify and recognize ground elements at different levels, analyze their attributes, and utilize class dependence and spatial relationships from a high-level perspective.

Pioneering work~\cite{shi2017can} proposed a solution to address the challenge of different semantics of the same ground elements in RS images under different geographical scales. They introduced a fully convolutional network (FCN) to generate comprehensive and robust sentences for RS images obtained from Google Earth and GaoFen-2, while ensuring desirable speed performance. They first tackled three subtasks, including key instance detection, environment analysis, and landscape analysis, and then integrated the results from each of these stages to generate language descriptions. Similarly, Lu et al.~\cite{lu2017exploring} provided instructions on how to comprehensively describe RS images, taking into account the scale ambiguity, category ambiguity, and rotation ambiguity, and created a large-scale image dataset for RS captioning by collecting more than ten thousand RS images. In contrast, Zhang et al.~\cite{zhang2017natural} employed a convolutional neural network to detect the primary objects in RS images, and a recurrent neural network language model to generate natural language descriptions of the detected objects.

Recent works also explored using the attention mechanism to better align visual features with text embeddings. For example, Zhang et al.~\cite{zhang2019vaa} proposed a visual aligning attention model that employs a well-crafted visual aligning loss. This loss was determined by explicitly evaluating the feature similarity between the attended image features and corresponding word embedding vectors. To address the effect of non-visual words in training the attention layer, they introduced a visual vocab that eliminates such words in sentences during the computation of the visual aligning loss. Additionally, to bridge the semantic gap between low-level features and high-level semantics in RS images, Zhang et al.~\cite{zhang2019description} utilized the Fully Convolutional Network (FCN) to generate image features and the attention mechanism to obtain intermediate vectors, which are then used as inputs to the LSTM decoder to produce descriptions of RS images. Following this work, Li et al.~\cite{li2020multi} proposed a multi-level attention model that contains three attention structures: attention to different areas of the image, attention to different words, and attention to vision and semantics. Moreover, they also corrected inaccuracies in existing datasets, including word errors, grammatical errors, and inappropriate captions. In another study, Shen et al.~\cite{shen2020remote} proposed a Two-stage Multi-task Learning Model based on the Variational Autoencoder (VAE) and Reinforcement Learning for RSIC. They first fine-tuned the CNN jointly with the VAE in the first stage and then utilized Transformer and Reinforcement Learning to both spatial and semantic features to generate the text description. Similarly, Kandala et al.~\cite{Kandala2022} employed a Transformer-based encoder–decoder network for RSIC. In particular, to deal with the limited training data, an auxiliary decoder, trained for multilabel scene classification, has been used to assist the encoder in the training process by leveraging its conceptual similarity to image captioning and its ability to highlight semantic classes. Instead of using the encoder-decoder architecture with a lack of explainability, Wang et al.~\cite{wang2020word} proposed an explainable word–sentence framework for RSIC, which consisted of two networks, a word extractor, and a sentence generator. The first network extracted the valuable words in the given RS image, and the second network organized these words into a coherent sentence. Besides, Li et al.~\cite{li2020truncation} proposed a novel truncation cross-entropy loss to solve the overfitting problem caused by a cross-entropy loss in RSIC. RSGPT~\cite{hu2023rsgpt} finetuned a single projection layer to align visual features with LLMs and showed impressive performance on both image captioning and visual question answering. Fig. \ref{fig_rsi_captioning} gives an overview of the RSGPT method. 

To address the issue of disregarding domain knowledge in previous methods, Zhao et al.~\cite{zhao2021high} proposed a fine-grained and structured attention-based model to utilize the structural characteristics of semantic contents in high-resolution RS images. In addition, Zia et al.~\cite{zia2022transforming} first employed the multi-scale visual feature encoder to extract detailed information from RS images and subsequently utilized the adaptive multi-head attention decoder to refine the description generation based on the extracted multi-scale features. Besides, they incorporated topic-sensitive word embedding to produce more human-like and innovative descriptions. More recently, Wei et al.~\cite{wei2023vlca} introduced a vision-language aligning network that jointly represents both vision and language for RSIC. This work was conducted on the DIOR-Captions dataset, a newly created dataset for enriching object detection in optical RS images dataset with manually annotated Chinese and English contents. 

We summarize the performance of comparison methods on the UCM-caption \cite{qu2016deep}, Sydney-caption \cite{qu2016deep}, and RSICD \cite{lu2017exploring} datasets in Table \ref{tab_image_captioning}. We report four evaluation metrics, including BLEU~\cite{papineni2002bleu}, METEOR~\cite{banerjee2005meteor}, ROUGE-L~\cite{chin2004rouge}, and CIDEr-D~\cite{vedantam2015cider}, to validate the effectiveness of comparing models. The results show that these existing visual-language models have achieved decent performance. Specifically, RSGPT~\cite{hu2023rsgpt} outperformed all compared methods, achieving the best performance.

However, it is noteworthy that the aforementioned three datasets, commonly employed for training visual-language models for remote sensing image captioning, comprise 2100, 613, and 10921 images, respectively. These quantities are significantly fewer than the millions of images typically employed in the training of visual-language models in the field of computer vision. This difference in dataset size does somewhat limit the potential for major improvements in the accuracy of visual-language models for remote sensing image captioning. More importantly, with the exception of the RSICap dataset introduced in RSGPT~\cite{hu2023rsgpt}, existing remote sensing caption datasets comprise only short captions. There is a significant need for large-scale datasets containing detailed captions to further advance the development of the remote sensing image captioning task.

\begin{table*}[!t]
    \centering
    % \resizebox{0.5\textwidth}{!}{
    \begin{tabular}{cccccccc}
         \toprule
         \multicolumn{7}{c}{UCM-Caption \cite{qu2016deep}} \\
         \midrule
         Method & BLEU-1 & BLEU-2 & BLEU-3 & BLEU-4 & METEOR & ROUGE\_L & CIDEr \\
         \midrule
         VLAD + RNN \cite{lu2017exploring} & 63.11 & 51.93 & 46.06 & 42.09 & 29.71 & 58.78 & 200.66 \\
         VLAD + LSTM \cite{lu2017exploring} & 70.16 & 60.85 & 54.96 & 50.30 & 34.64 & 65.20 & 231.31 \\
         mRNN \cite{qu2016deep} & 60.10 & 50.70 & 32.80 & 20.80 & 19.30 & - & 214.00 \\
         mLSTM \cite{qu2016deep} & 63.50 & 53.20 & 37.50 & 21.30 & 20.30 & - & 222.50 \\
         mGRU \cite{mgru} & 42.56 & 29.99 & 22.91 & 17.98 & 19.41 & 37.97 & 124.82 \\
         mGRU embedword \cite{mgru} & 75.74 & 69.83 & 64.51 & 59.98 & 36.85 & 66.74 & 279.24 \\
         CSMLF \cite{csmlf} & 37.71 & 14.85 & 7.63 & 5.05 & 9.44 & 29.86 & 13.51 \\
         SAA \cite{sound} & 79.62 & 74.01 & 69.09 & 64.77 & 38.59 & 69.42 & 294.51 \\
         Soft-attention \cite{softattention} & 74.54 & 65.45 & 58.55 & 52.50 & 38.86 & 72.37 & 261.24 \\
         Hard-attention \cite{softattention} & 81.57 & 73.12 & 67.02 & 61.82 & \textbf{42.63} & 76.98 & 299.47 \\
         SD-RSIC \cite{sd-rsic} & 74.80 & 66.40 & 59.80 & 53.80 & 39.00 & 69.50 & 213.20 \\
         RTRMN (semantic) \cite{rtrmn} & 55.26 & 45.15 & 39.62 & 35.87 & 25.98 & 55.38 & 180.25 \\
         RTRMN (statistical) \cite{rtrmn} & 80.28 & 73.22 & 68.21 & 63.93 & 42.58 & 77.26 & 312.70 \\
         SVM-D BOW \cite{svm-d} & 76.35 & 66.64 & 58.69 & 51.95 & 36.54 & 68.01 & 271.42 \\
         SVM-D CONC \cite{svm-d} & 76.53 & 69.47 & 64.17 & 59.42 & 37.02 & 68.77 & 292.28 \\
         Post-processing \cite{post-processing} & 79.73 & 72.98 & 67.44 & 62.62 & 40.80 & 74.06 & 309.64 \\
         RSGPT \cite{hu2023rsgpt} & \textbf{86.12} & \textbf{79.14} & \textbf{72.31} & \textbf{65.74} & 42.21 & \textbf{78.34} & \textbf{333.23} \\
         \midrule
         \multicolumn{7}{c}{Sydney-captions \cite{qu2016deep}} \\
         \midrule
         VLAD + RNN \cite{lu2017exploring} & 56.58 & 45.14 & 38.07 & 32.79 & 26.72 & 52.71 & 93.72 \\
         VLAD + LSTM \cite{lu2017exploring} & 49.13 & 34.12 & 27.60 & 23.14 & 19.30 & 42.01 & 91.64 \\
         mRNN \cite{qu2016deep} & 51.30 & 37.50 & 20.40 & 19.30 & 18.50 & - & 161.00 \\
         mLSTM \cite{qu2016deep} & 54.60 & 39.50 & 22.30 & 21.20 & 20.50 & - & 186.00 \\
         mGRU \cite{mgru} & 69.64 & 60.92 & 52.39 & 44.21 & 31.12 & 59.17 & 171.55 \\
         mGRU embedword \cite{mgru} & 68.85 & 60.03 & 51.81 & 44.29 & 30.36 & 57.47 & 168.94 \\
         CSMLF \cite{csmlf} & 59.98 & 45.83 & 38.69 & 34.33 & 24.75 & 50.18 & 75.55 \\
         SAA \cite{sound} & 68.82 & 60.73 & 52.94 & 45.39 & 30.49 & 58.20 & 170.52 \\
         Soft-attention \cite{softattention} & 73.22 & 66.74 & 62.23 & 58.20 & 39.42 & 71.27 & 249.93 \\
         Hard-attention \cite{softattention} & 75.91 & 66.10 & 58.89 & 52.58 & 38.98 & 71.89 & 218.19 \\
         SD-RSIC \cite{sd-rsic} & 72.40 & 62.10 & 53.20 & 45.10 & 34.20 & 63.60 & 139.50 \\
         SVM-D BOW \cite{svm-d} & 77.87 & 68.35 & 60.23 & 53.05 & 37.97 & 69.92 & 227.22 \\
         SVM-D CONC \cite{svm-d} & 75.47 & 67.11 & 59.70 & 53.08 & 36.43 & 67.46 & 222.22 \\
         Post-processing \cite{post-processing} & 78.37 & 69.85 & 63.22 & 57.17 & 39.49 & 71.06 & 255.53 \\
         RSGPT \cite{hu2023rsgpt} & \textbf{82.26} & \textbf{75.28} & \textbf{68.57} & \textbf{62.23} & \textbf{41.37} & \textbf{74.77} & \textbf{273.08} \\
         \midrule
         \multicolumn{7}{c}{RSICD \cite{lu2017exploring}} \\
         \midrule
         VLAD + RNN \cite{lu2017exploring} & 49.38 & 30.91 & 22.09 & 16.77 & 19.96 & 42.42 & 103.92 \\
         VLAD + LSTM \cite{lu2017exploring} & 50.04 & 31.95 & 23.19 & 17.78 & 20.46 & 43.34 & 118.01 \\
         mRNN \cite{qu2016deep} & 45.58 & 28.25 & 18.09 & 12.13 & 15.69 & 31.26 & 19.15 \\
         mLSTM \cite{qu2016deep} & 50.57 & 32.42 & 23.29 & 17.46 & 17.84 & 35.02 & 31.61 \\
         mGRU \cite{mgru} & 42.56 &29.99 & 22.91 & 17.98 & 19.41 & 37.97 & 124.82 \\
         mGRU embedword \cite{mgru} & 60.94 & 46.24 & 36.80 & 29.81 & 26.14 & 48.20 & \textbf{159.54} \\
         CSMLF \cite{csmlf} & 57.59 & 38.59 & 28.32 & 22.17 & 21.28 & 44.55 & 52.97 \\
         SAA \cite{sound} & 59.35 & 45.11 & 35.29 & 28.08 & 26.11 & 49.57 & 132.35 \\
         Soft-attention \cite{softattention} & 65.13 & 49.04 & 39.00 & 32.30 & 26.39 & 49.69 & 90.58 \\
         SD-RSIC \cite{sd-rsic} & 64.50 & 47.10 & 36.40 & 29.40 & 24.90 & 51.90 & 77.50 \\
         RTRMN (semantic) \cite{rtrmn} & 62.01 & 46.23 & 36.44 & 29.71 & 28.29 & \textbf{55.39} & 151.46 \\
         RTRMN (statistical) \cite{rtrmn} & 61.02 & 45.14 & 35.35 & 28.59 & 27.51 & 54.52 & 148.20 \\
         SVM-D BOW \cite{svm-d} & 61.12 & 42.77 & 31.53 & 24.11 & 23.03 & 45.88 & 68.25 \\
         SVM-D CONC \cite{svm-d} & 59.99 & 43.47 & 33.55 & 26.89 & 22.99 & 45.57 & 68.54 \\
         MLAT \cite{mlat} & 66.90 & 51.13 & 41.14 & 34.21 & 27.31 & 50.57 & 94.27 \\
         Post-processing \cite{post-processing} & 62.90 & 45.99 & 35.68 & 28.68 & 25.30 & 47.34 & 75.56 \\
         RSGPT \cite{hu2023rsgpt} & \textbf{70.32} & \textbf{54.23} & \textbf{44.02} & \textbf{36.83} & \textbf{30.10}  & 53.34 & 102.94 \\
         \bottomrule
    \end{tabular}
    % }
    \caption{Remote sensing image captioning performance on the UCM-caption, Sydney-caption, and RSICD datasets. Numbers are borrowed from \cite{hu2023rsgpt}}
    \label{tab_image_captioning}
\end{table*}

 \begin{figure*}[ht]
     \centering
     \includegraphics[width=18cm]{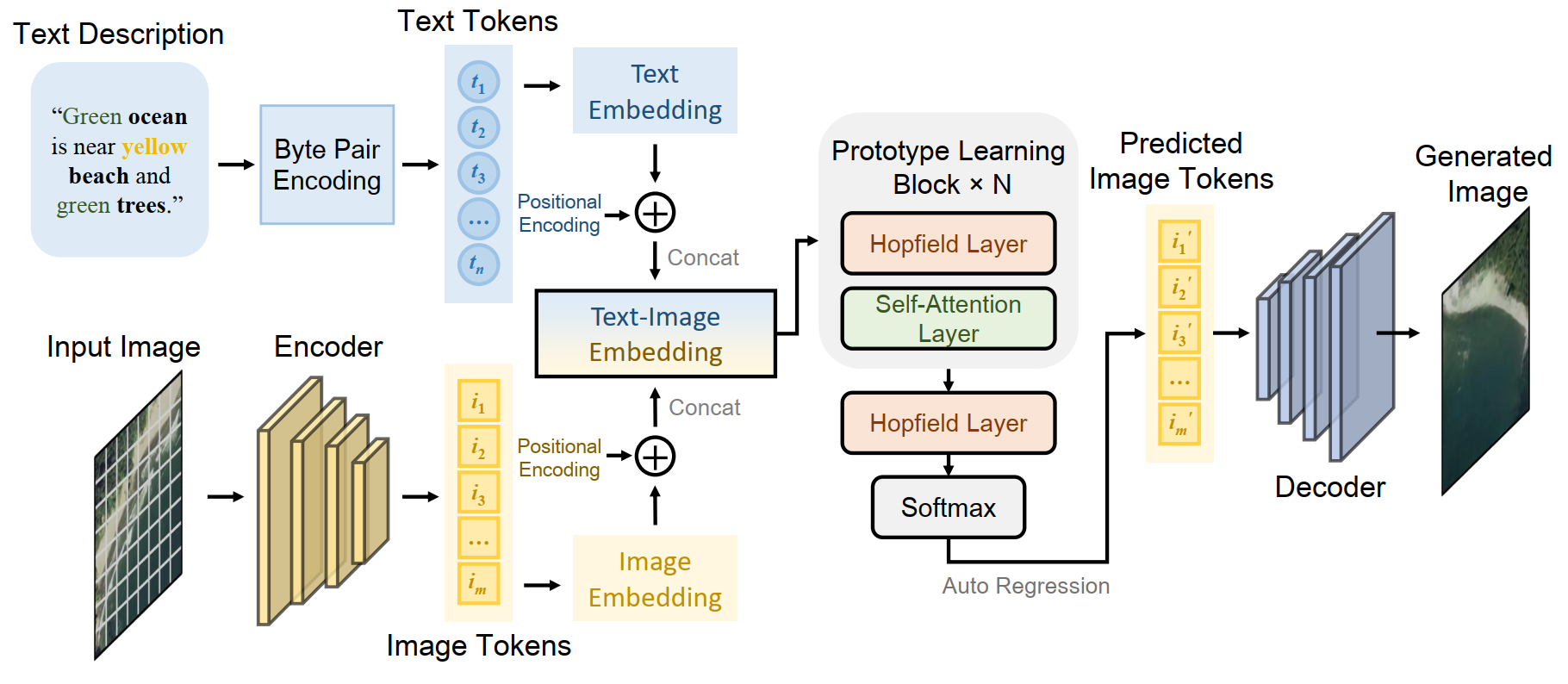}
     \caption{Illustration of the proposed method in~\cite{xu2022txt2img} for text-based remote sensing image generation.}
     \label{fig_T2IG}
 \end{figure*}
 
\subsection{Text-based Image Generation}

Text-based image generation is an emerging field of research that combines natural language processing and computer vision to create realistic images from textual descriptions. The application of this technology to remote sensing (RS) images holds significant potential in real-world applications. One area where it could be beneficial is in assisting urban planners by generating realistic RS images based on their text descriptions. This would enable them to evaluate the feasibility of their designs and make better-informed decisions. Another potential use case is in generating high-quality labeled datasets of RS images, which is often a challenging and time-consuming process. Text-based image generation techniques could be used to create synthetic RS datasets from text descriptions, alleviating the shortage of labeled samples.

Several research studies have investigated text-based RS image generation using generative adversarial networks (GANs). For example, Bejiga et al.~\cite{bejiga2019retro} conducted a pioneering investigation into the synthesis of RS images using text descriptions by utilizing GANs. They generated RS images based on ancient text descriptions of geographical areas by transforming text representation into pixel values that captured image characteristics such as size and color. Besides, Chen et al.~\cite{chen2021remote} proposed an enhanced GAN, called Text-based Deeply-supervised GAN (BTD-sGAN), to address the low quantity and poor quality of RS images. BTD-sGAN employed a two-layer UNet++~\cite{zhou2018unet++} framework and took Perlin Noise, image segmentation graph, and encoded text vector as input to generate RS images. They realized an Inception Score of 3.66 on the natural image dataset Oxford-102 and demonstrated promising qualitative results on their own remote sensing dataset. Moreover, Zhao et al.~\cite{zhao2021text} introduced a structured GAN, i.e., StrucGAN, to generate RS images based on textual descriptions while considering structural information. The proposed StrucGAN utilized a bidirectional long short-term memory network(LSTM~\cite{hochreiter1997long}) as the text encoder to capture semantic features and adopted the StackGANs to generate images gradually from small to large scales. The model also included a branch consisting of a region proposal module (RPM) and a discriminator to ensure the generation of structurally reasonable images. This method achieved an Inception Score of 5.84 on the popular RSICD dataset and a top-10 precision of 16.20\%. Furthermore, Xu et al.~\cite{xu2022txt2img} presented a novel approach, Txt2Img-MHN, to generate RS images from text descriptions using a modern Hopfield network~\cite{ramsauer2020hopfield}. As illustrated in Fig.~\ref{fig_T2IG}, unlike the previous work focused on learning joint text-image representations, the Txt2Img-MHN applied the Hopfield network to the text-image embeddings and performed prototype learning hierarchically, which guarantees better coarseness and finesse in the learning steps along with richer semantic encoding. They attained notable performance metrics on the RSICD test dataset, including an Inception Score of 5.99 and an FID Score of 102.44. In addition, we summarize the Inception Score~\cite{salimans2016improved} and FID Score~\cite{heusel2017gans} of all comparison methods on the RSICD test dataset in Table \ref{tab_ig}.

\begin{table}[]
\centering
\begin{tabular}{ccc}
\toprule
Method              & Inception Score($\uparrow$) & FID Score($\downarrow$) \\ \midrule
Attn-GAN~\cite{xu2018attngan}        & \textbf{11.71}           & 95.81     \\
DAE-GAN~\cite{ruan2021dae}             & 7.71            & 93.15     \\
DF-GAN~\cite{tao2022df}              & 9.51            & 109.41    \\
Lafite~\cite{zhou2022towards}              & 10.7            & \textbf{74.11}     \\
DALL-E~\cite{ramesh2021zero}              & 2.59            & 191.93    \\
Txt2Img-MHN (VQVAE)~\cite{ramsauer2020hopfield} & 3.51            & 175.36    \\
Txt2Img-MHN (VQGAN)~\cite{ramsauer2020hopfield} & 5.99            & 102.44    \\ \bottomrule
\end{tabular}
\caption{Text-based Remote Sensing Image Generation on the RSICD test set. Numbers are borrowed from~\cite{xu2022txt2img}}
\label{tab_ig}
\end{table}

\begin{figure*}[ht]
     \centering
     \includegraphics[width=18cm]{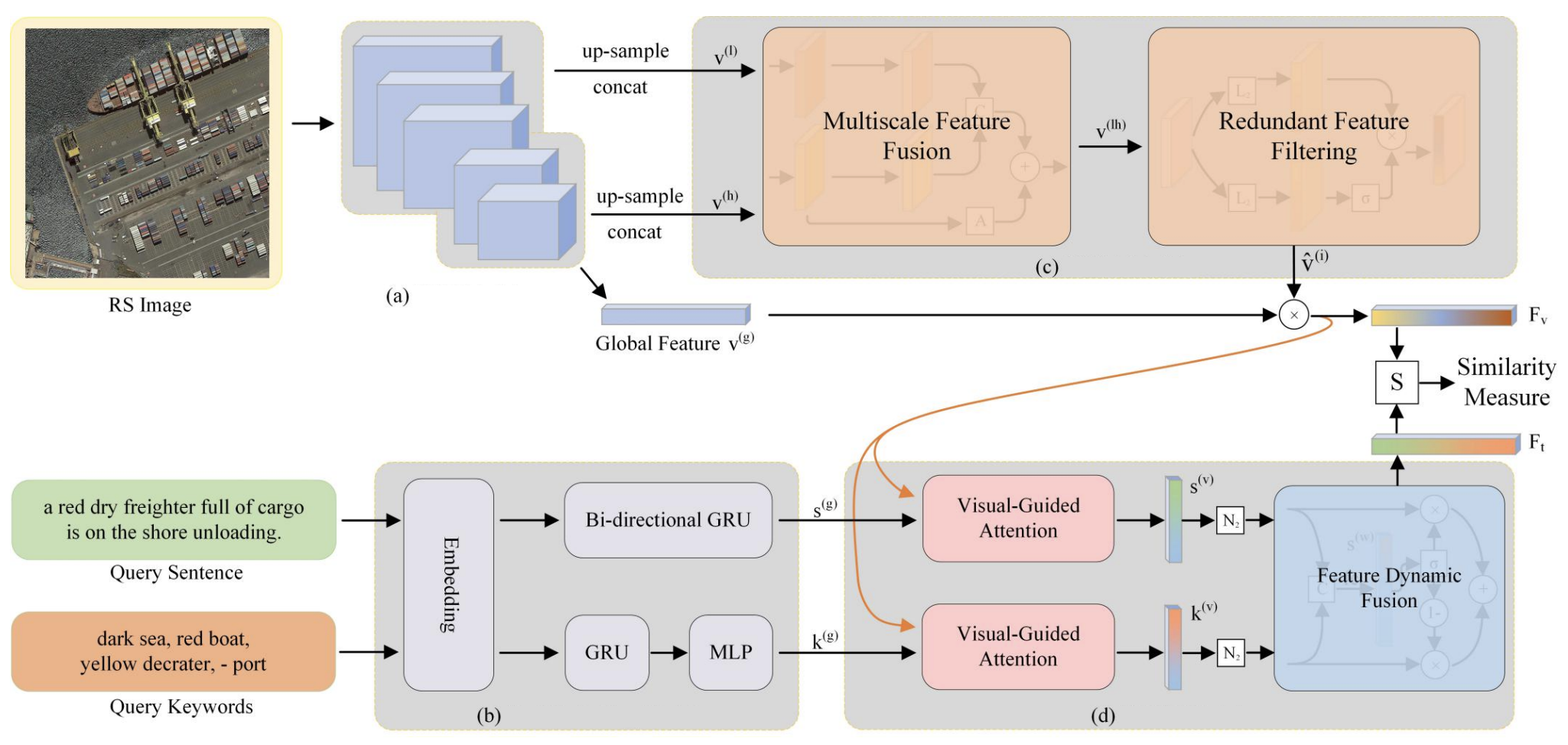}
     \caption{Illustration of the proposed method in~\cite{yuan2022exploring} for text-based remote sensing image retrieval.}
     \label{fig_T2IR}
 \end{figure*}
 
\subsection{Text-based Image Retrieval}
The efficient organization and management of vast amounts of remote sensing (RS) data has long posed a significant challenge to the RS community. To address this challenge, text-based image retrieval (TBIR) has emerged as a prominent research topic, aiming to provide an effective solution for RS data management. The primary objective of image retrieval is to extract a specific image from a large dataset, and it has gained considerable attention recently. The fundamental idea is to narrow down the search for the targeted image and retrieve the image that matches a particular query. This task is valuable in practical applications such as deforestation detection, visual navigation, and urban planning.

To achieve this, Abdullah et al.~\cite{abdullah2020textrs} constructed a new dataset, named TextRS, for the text-based image retrieval task. The dataset comprised images from four distinct scene datasets, each annotated with five corresponding sentences. To retrieve the image that matches a particular query, they also proposed a novel Deep Bidirectional Triplet Network composed of LSTM~\cite{hochreiter1997long} and CNNs for RS retrieval by matching text to image representations. The experimental results demonstrated the model's promising performance in terms of Recall@K, achieving values of 7.20\%, 51.39\%, and 73.02\% for K = 1, 5, and 10, respectively. Similarly, Rahhal et al.~\cite{rahhal2020deep} proposed an unsupervised text-image retrieval model for RS images by utilizing a visual Big Transfer Model to learn image representation and a bidirectional LSTM~\cite{hochreiter1997long} network to encode textual descriptions. The model was optimized using an unsupervised embedding loss to ensure that an image's features are closest to its corresponding textual description, and vice versa, while being dissimilar to other image features. This study accomplished text-to-image retrieval on the TextRS dataset with the following scores: 19.02\%, 55.25\%, and 71.72\% for R@1, R@5, and R@10, respectively. Moreover, on the Merced dataset, the achieved scores for R@1, R@5, and R@10 are 21.86\%, 60.00\%, and 75.58\%, respectively.

% {\color{red}Missing summary sentence at the beginning.}
Yuan et al.~\cite{yuan2022remote} proposed a novel remote sensing text-image retrieval framework that integrated local and global information, using a dynamic fusion module to effectively integrate features from different levels. In response to the problem of existing methods that were limited to processing queries formulated in English, Rahhal et al.~\cite{al2022multilanguage} proposed a multi-language framework composed of a language encoder for generating language representation features from the textual description and a vision encoder for extracting visual features from the corresponding image. The authors trained the model on text descriptions from four different languages, including English, Arabic, French, and Italian. In addition, Chen et al.~\cite{cheng2021deep} introduced a novel cross-modal image-text retrieval network based on a designed semantic alignment module, which was used to obtain more discriminative feature representations by using attention and gate mechanisms to establish the direct relationship between RS images and their paired text data.

Yuan et al. \cite{yuan2021lightweight} proposed a simple but effective cross-modal model for RS image retrieval to address the challenges posed by the multi-scale and target redundancy characteristics of RS data. The proposed model incorporated multi-scale information and dynamically filtered out redundant features during RS image encoding, while text features were obtained via lightweight group convolution. Furthermore, to improve retrieval performance, they developed a novel hidden supervised optimization method based on knowledge distillation. This method enabled the proposed model to acquire dark knowledge of the multi-level layers and representation layers in the teacher network, which significantly improved the accuracy of our lightweight model. In a similar vein, Yuan et al.~\cite{yuan2022exploring} employed a multi-scale self-attention module to obtain the salient features of RS images to filter redundant features dynamically, which is illustrated in Fig.~\ref{fig_T2IR}. The authors also constructed a fine-grained RS Image-Text Match dataset, which enabled RS image retrieval based on keywords and sentences both independently. Across four datasets, including Sydney, UCM, RSICD, and RSITMD, the method mentioned above achieved mean recall scores of approximately 52\%, 46\%, 30\%, and 16\%. Recently, Rahhal et al.~\cite{rahhal2023contrasting} proposed an efficient text–image retrieval approach that utilized vision and language transformers to align the visual representations of RS images to their corresponding textual representations using two contrastive losses for image-to-text and text-to-image classification. This method provided a slight improvement in retrieval performance across datasets compared to the previous method.

Furthermore, we provide a summary of the performance of state-of-the-art methods on RSICD, RSITMD, UCM, and Sydney datasets in Table \ref{tab_ir}. In the table presented, we disclose the scores for R@K, calculated for K values of 1, 5, and 10, as well as the mR scores. Here, R@K reflects the proportion of ground truth found within the top K results, while mR represents the average of the six recall rates for R@K, as introduced by Huang et al.~\cite{huang2018learning}.

\begin{table*}[]
\begin{tabular}{c|ccc|ccc|c|ccc|ccc|c}
\toprule \multirow{3}{*}{ Approach } & \multicolumn{7}{c|}{ RSICD dataset } & \multicolumn{7}{c}{ RSITMD dataset } \\
\cline{2-15}
 & \multicolumn{3}{c|}{ Sentence Retrieval } & \multicolumn{3}{c|}{ Image Retrieval } & \multirow{2}{*}{\(\mathrm{mR}\)} & \multicolumn{3}{c|}{ Sentence Retrieval } & \multicolumn{3}{c|}{ Image Retrieval } & \multirow{2}{*}{\(\mathrm{mR}\)} \\
 \cline{2-7}\cline{9-14}
 & \(\mathrm{R} @ 1\) & R@5 & R@10 & \(\mathrm{R} @ 1\) & \(\mathrm{R} @ 5\) & R@10 & & R@1 & R@5 & R@10 & \(\mathrm{R} @ 1\) & \(\mathrm{R} @ 5\) & R@10 & \\
 \hline
 VSE++~\cite{faghri2017vse++} & 3.38 & 9.51 & 17.46 & 2.82 & 11.32 & 18.10 & 10.43 & 10.38 & 27.65 & 39.60 & 7.79 & 24.87 & 38.67 & 24.83 \\
 SCAN-t2i~\cite{lee2018stacked} & 4.39 & 10.90 & 17.64 & 3.91 & 16.20 & 26.49 & 13.25 & 10.18 & 28.53 & 38.49 & 10.10 & 28.98 & 43.53 & 26.64 \\
 SCAN-i2t~\cite{lee2018stacked} & \textbf{5.85} & 12.89 & 19.84 & 3.71 & 16.40 & 26.73 & 14.23 & 11.06 & 25.88 & 39.38 & 9.82 & 29.38 & 42.12 & 26.28 \\
 CAMP-triplet~\cite{wang2019camp} & 5.12 & 12.89 & 21.12 & 4.15 & 15.23 & 27.81 & 14.39 & \textbf{11.73} & 26.99 & 38.05 & 8.27 & 27.79 & 44.34 & 26.20 \\
 CAMP-bce~\cite{wang2019camp} & 4.20 & 10.24 & 15.45 & 2.72 & 12.76 & 22.89 & 11.38 & 9.07 & 23.01 & 33.19 & 5.22 & 23.32 & 38.36 & 22.03 \\
 MTFN~\cite{wang2019matching} & 5.02 & 12.52 & 19.74 & 4.90 & 17.17 & 29.49 & 14.81 & 10.40 & 27.65 & 36.28 & 9.96 & 31.37 & 45.84 & 26.92 \\
 AMFMN-soft~\cite{yuan2022exploring} & 5.05 & 14.53 & 21.57 & \textbf{5.05} & \textbf{19.74} & 31.04 & 16.02 & 11.06 & 25.88 & 39.82 & 9.82 & 33.94 & 51.90 & 28.74 \\
 AMFMN-fusion~\cite{yuan2022exploring} & 5.39 & \textbf{15.08} & \textbf{23.40} & 4.90 & 18.28 & \textbf{31.44} & \textbf{16.42} & 11.06 & \textbf{29.20} & 38.72 & 9.96 & 34.03 & 52.96 & 29.32 \\
 \multirow[t]{2}{*}{ AMFMN-sim } & 5.21 & 14.72 & 21.57 & 4.08 & 17.00 & 30.60 & 15.53 & 10.63 & 24.78 & \textbf{41.81} & \textbf{11.51} & \textbf{34.69} & \textbf{54.87} & \textbf{29.72} \\
 \hline
  \multirow{3}{*}{ Approach } & \multicolumn{7}{c|}{ UCM dataset } & \multicolumn{6}{c}{ Sydney dataset } & \\
  \cline{2-15}
  & \multicolumn{3}{c|}{ Sentence Retrieval } & \multicolumn{3}{c|}{ Image Retrieval } & \multirow{2}{*}{\(\mathrm{mR}\)} & \multicolumn{3}{c|}{ Sentence Retrieval } & \multicolumn{3}{c|}{ Image Retrieval } & \multirow{2}{*}{\(\mathrm{mR}\)} \\
  \cline{2-7}\cline{9-14}
 & \(\mathrm{R} @ 1\) & R@5 & \(\mathrm{R} @ 10\) & \(\mathrm{R} @ 1\) & \(\mathrm{R} @ 5\) & \(\mathrm{R} @ 10\) & & \(\mathrm{R} @ 1\) & \(\mathrm{R} @ 5\) & \(\mathrm{R} @ 10\) & \(\mathrm{R} @ 1\) & \(\mathrm{R} @ 5\) & \(\mathrm{R} @ 10\) & \\
 \hline
 VSE++~\cite{faghri2017vse++} & 12.38 & 44.76 & 65.71 & 10.10 & 31.80 & 56.85 & 36.93 & 24.14 & 53.45 & 67.24 & 6.21 & 33.56 & 51.03 & 39.27 \\
 SCAN-t2i~\cite{lee2018stacked} & 14.29 & 45.71 & 67.62 & 12.76 & 50.38 & 77.24 & 44.67 & 18.97 & 51.72 & 74.14 & \textbf{17.59} & 56.90 & 76.21 & 49.26 \\
 SCAN-i2t~\cite{lee2018stacked} & 12.85 & 47.14 & \textbf{69.52} & 12.48 & 46.86 & 71.71 & 43.43 & 20.69 & 55.17 & 67.24 & 15.52 & 57.59 & 76.21 & 48.74 \\
 CAMP-triplet~\cite{wang2019camp} & 10.95 & 44.29 & 65.71 & 9.90 & 46.19 & 76.29 & 42.22 & 20.69 & 53.45 & \textbf{75.86} & 14.14 & 42.07 & 69.66 & 45.98 \\
 CAMP-bce~\cite{wang2019camp} & 14.76 & 46.19 & 67.62 & 11.71 & 47.24 & 76.00 & 43.92 & 15.52 & 51.72 & 72.41 & 11.38 & 51.72 & 76.21 & 46.49 \\
 MTFN~\cite{wang2019matching} & 10.47 & 47.62 & 64.29 & \textbf{14.19} & 52.38 & 78.95 & 44.65 & 20.69 & 51.72 & 68.97 & 13.79 & 55.51 & 77.59 & 48.05 \\
 AMFMN~\cite{yuan2022exploring} & 12.86 & \textbf{51.90} & 66.67 & \textbf{14.19} & 51.71 & 78.48 & 45.97 & 20.69 & 51.72 & 74.14 & 15.17 & 58.62 & 80.00 & 50.06 \\
 AMFMN-fusion~\cite{yuan2022exploring} & \textbf{16.67} & 45.71 & 68.57 & 12.86 & \textbf{53.24} & \textbf{79.43} & \textbf{46.08} & 24.14 & 51.72 & \textbf{75.86} & 14.83 & 56.55 & 77.89 & 50.17 \\
 AMFMN-sim~\cite{yuan2022exploring} & 14.76 & 49.52 & 68.10 & 13.43 & 51.81 & 76.48 & 45.68 & \textbf{29.31} & \textbf{58.62} & 67.24 & 13.45 & \textbf{60.00} & \textbf{81.72} & \textbf{51.72} \\
\bottomrule
\end{tabular}
\caption{Results of SENTENCE-based and IMAGE-based 
 Remote Sensing RETRIEVAL ON RSICD, RSITMD, UCM, AND SYDNEY Datasets. Numbers are borrowed from~\cite{yuan2022exploring}.}
\label{tab_ir}
\end{table*}

 \begin{figure*}[ht]
     \centering
     \includegraphics[width=16cm]{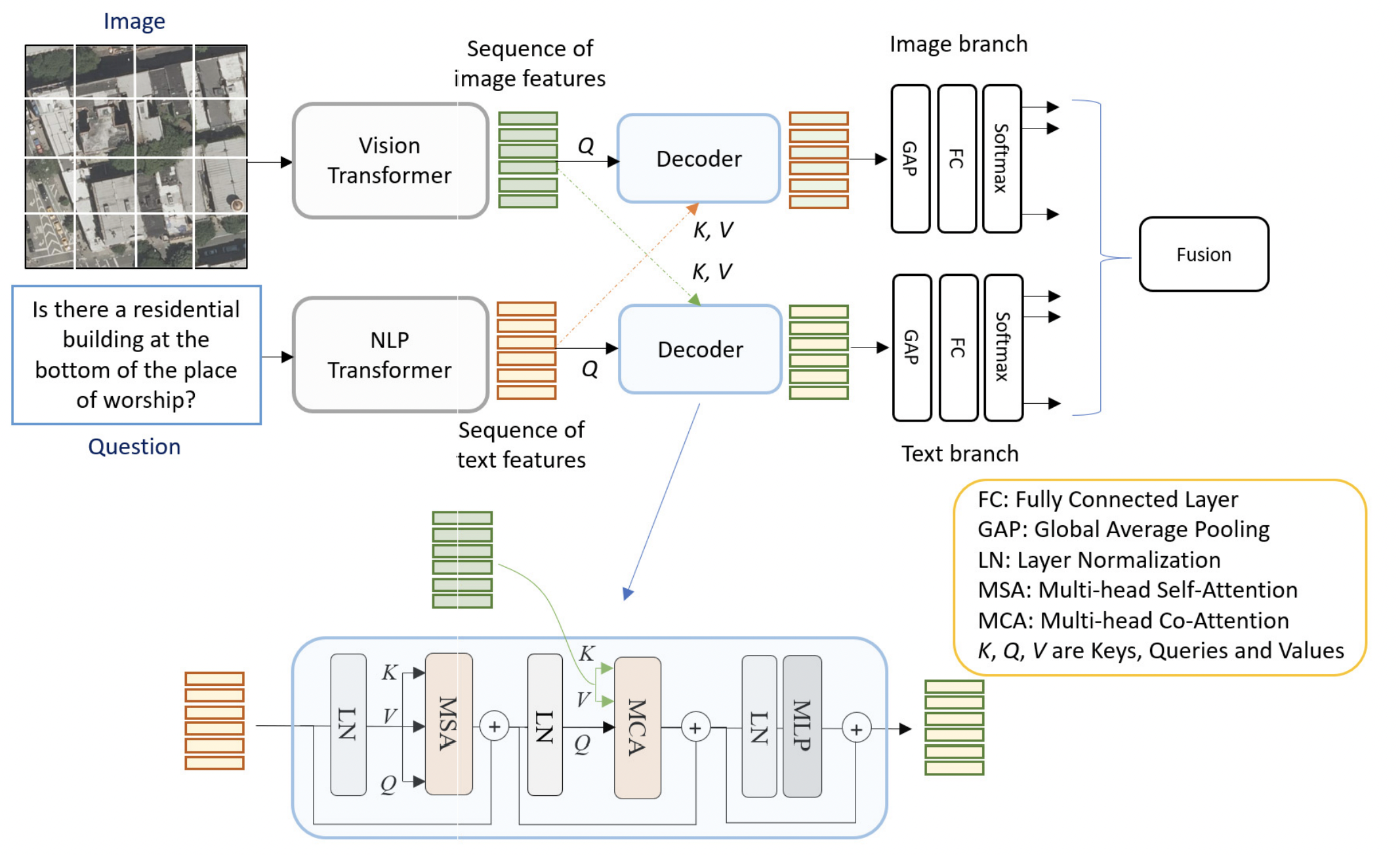}
     \caption{Framework of a remote sensing visual question answering method proposed in~\cite{bazi2022bi}. The method uses a vision-language transformer to encode image and question pairs into textual and visual feature representations.}
     \label{fig_vqa}
 \end{figure*}
 
\subsection{Visual Question Answering}
Visual question answering (VQA) is a task that seeks to provide answers to image-related questions. While it has gained popularity in the field of computer vision, it is still in its early stages in the remote sensing (RS) community. The remote sensing VQA system enables non-expert users to interact with RS images using natural language questions as queries, thus enabling a user-friendly and high-level understanding of RS images. Pioneering work~\cite{lobry2020rsvqa} built the first large-scale VQA benchmark dataset for RS images, called RSVQA. Both low- and high-resolution RS images were collected data from OpenStreetMap, with human-generated questions and answers relevant to the image. In total, there are 772 images with 77,232 question-answer pairs in the low-resolution set, 
10,659 images with 1,066,316 question-answer pairs in the high-resolution set. In~\cite{lobry2020rsvqa}, the authors provided a benchmark method that uses Convolutional neural networks (CNNs) for visual feature learning, and LSTM~\cite{hochreiter1997long} networks were adopted for text embedding extraction. Mutual attention was further designed to enhance the alignment between visual and textual features. The proposed method achieved an overall accuracy close to 80\% on both low- and high-resolution sets. Zheng et al.~\cite{zheng2021mutual} introduced a remote sensing image visual question answering dataset called RSIVQA that contains around 37k images. They also developed a mutual attention network to exploit semantic correspondence between visual and textual features, with a bilinear module to conduct feature fusion. The proposed method achieved an overall accuracy of 77\% on their RSIVQA dataset. Chappuis et al.~\cite{chappuis2022language} proposed to use large language transformers, e.g. BERT~\cite{devlin2018bert}, to learn textural features and demonstrated better performance than recurrent neural networks. This method obtained an overall accuracy close to 84\% on the low-resolution set of the RSVQA dataset. Recently Yuan et al~\cite{yuan2023multilingual} designed a multilingual version of the RSVQA dataset and improved the robustness of the RSVQA model with the augmented dataset.
 
Unlike previous methods that mainly focus on solving remote sensing VQA in a closed-ended scenario, Al et al.~\cite{al2022open} introduced a novel dataset, VQA-TextRS, with 2,144 RS images from 134 classes, where both questions and answers were created through human annotation. It incorporates a diverse range of open-ended question-answer pairs. In order to address the task of open-ended VQA, they utilized vision and language transformer networks to extract visual and textual features from both images and accompanying questions, followed by a transformer decoder that leverages a cross-attention mechanism to integrate the two modalities. As a result, the proposed method achieved an accuracy of 84.01\% in relation to queries about the presence of objects within the image. A similar idea was proposed in~\cite{bazi2022bi} where the authors proposed to use the CLIP~\cite{radford2021learning} model to embed images and questions as visual and textual representations, followed by an attention mechanism to learn correlations between these representations. Fig. \ref{fig_vqa} gives an overview of the proposed method for remote sensing VQA. The proposed method achieved an overall accuracy of 85\% on the low-resolution RSVQA dataset, and an overall accuracy of 85\% and 81\% on test set 1 and test set 2 of the high-resolution RSVQA dataset. In contrast to traditional approaches that utilize visual encoder networks for feature learning, Chappuis et al.~\cite{chappuis2022prompt} introduced a novel method that converts context information from images into text prompts that a language model can process. This enables the processing of both questions and visual contexts within a unified language model. Yuan et al.~\cite{yuan2022easy} introduced a self-paced curriculum learning (SPCL)-based training technique to train the RSVQA model in an easy-to-hard way. The proposed method obtained an overall accuracy of 83\% on the low-resolution RSVQA dataset, 84.2\% and 79.3\% on test set 1 and test set 2 of the high-resolution RSVQA dataset, and an overall accuracy of 79.7\% on the RSIVQA dataset. 

Recently, an interesting idea was proposed in~\cite{yuan2022change}. The authors leveraged a VQA system for change detection on multitemporal aerial images. They built a change detection-based visual question-answering (CDVQA) dataset that contains multitemporal images and question-answer pairs using an automatic question–answer generation method. This dataset contains 4,662 image pairs with semantic change maps. Furthermore, they developed a baseline method for the CDVQA task which contains four parts: multitemporal feature encoding, multitemporal fusion, multimodal fusion, and answer prediction.
The baseline method achieved an overall accuracy of 68.0\% and 63.4\% on test set 1 and test set 2 of the CDVQA dataset. 

While existing RSVQA methods have demonstrated promising performance, there is a notable disparity in their performance when it comes to object counting compared to object presence and classification tasks~\cite{yuan2022easy}. Unlike previous methods that primarily utilized pre-defined question inputs, recent work~\cite{bashmal2023visual} introduced the Visual Question Generation (VQG) approach, which is capable of generating intelligent, natural language questions about remote sensing (RS) images. We summarize the performance of comparison methods on the RSVQA-HR \cite{lobry2020rsvqa} and RSVQA-LR \cite{lobry2020rsvqa} datasets in Table \ref{tab_vqa_hr} and Table \ref{tab_vqa_lr}, respectively. We report the accuracy for each question type and the average accuracy for all types in the Tables. From these tables, RSGPT~\cite{hu2023rsgpt} outperformed all compared methods.

\begin{table}[t]
    \centering
    \begin{tabular}{cccccc}
         \toprule
         \multicolumn{4}{c}{RSVQA-HR Test Set 1 \cite{lobry2020rsvqa}} \\
         \midrule
         Method & Presence & Comparison & Average Accuracy \\
         \midrule
         RSVQA \cite{lobry2020rsvqa} & 90.43 & 88.19 & 89.31 \\
         EasyToHard \cite{yuan2022easy} & 91.39 & 89.75 & 90.57 \\
         Bi-Modal \cite{bazi2022bi} & 92.03 & 91.83 & 91.93 \\
         SHRNet \cite{shrnet} & \textbf{92.45} & 91.68 & \textbf{92.07} \\
         RSGPT \cite{hu2023rsgpt} & 91.86 & \textbf{92.15} & 92.00 \\
         \midrule
         \multicolumn{4}{c}{RSVQA-HR Test Set 2 \cite{lobry2020rsvqa}} \\
         \midrule
         RSVQA \cite{lobry2020rsvqa} & 86.26 & 85.94 & 86.10 \\
         EasyToHard \cite{yuan2022easy} & 87.97 & 87.68 & 87.83 \\
         Bi-Modal \cite{bazi2022bi} & 89.37 & 89.62 & 89.50 \\
         SHRNet \cite{shrnet} & 89.81 & 89.44 & 89.63 \\
         RSGPT \cite{hu2023rsgpt} & \textbf{89.87} & \textbf{89.68} & \textbf{89.78} \\
         \bottomrule
    \end{tabular}
    \caption{Remote sensing visual question answering performance on the RSVQA-HR Test Set 1 and RSVQA-HR Test Set 2. Numbers are borrowed from \cite{hu2023rsgpt}.}
    \label{tab_vqa_hr}
\end{table}

\begin{table}[t]
    \centering
    \resizebox{0.5\textwidth}{!}{
    \begin{tabular}{ccccccc}
         \toprule
         Method & Presence & Comparison & Rural/Urban & Average Accuracy \\
         \midrule
         RSVQA \cite{lobry2020rsvqa} & 87.46 & 81.50 & 90.00 & 86.32 \\
         EasyToHard \cite{yuan2022easy} & 90.66 & 87.49 & 91.67 &  89.94\\
         Bi-Modal \cite{bazi2022bi} & 91.06 & 91.16 & 92.66 & 91.63 \\
         SHRNet \cite{shrnet} & 91.03 & 90.48 & 94.00 & 91.84 \\
         RSGPT \cite{hu2023rsgpt} & \textbf{91.17} & \textbf{91.70} & \textbf{94.00} & \textbf{92.29} \\
         \bottomrule
    \end{tabular}
    }
    \caption{Remote sensing visual question answering performance on the RSVQA-LR \cite{lobry2020rsvqa} dataset. Numbers are borrowed from \cite{hu2023rsgpt}.}
    \label{tab_vqa_lr}
\end{table}

\begin{figure}[ht]
     \centering
     \includegraphics[width=8cm]{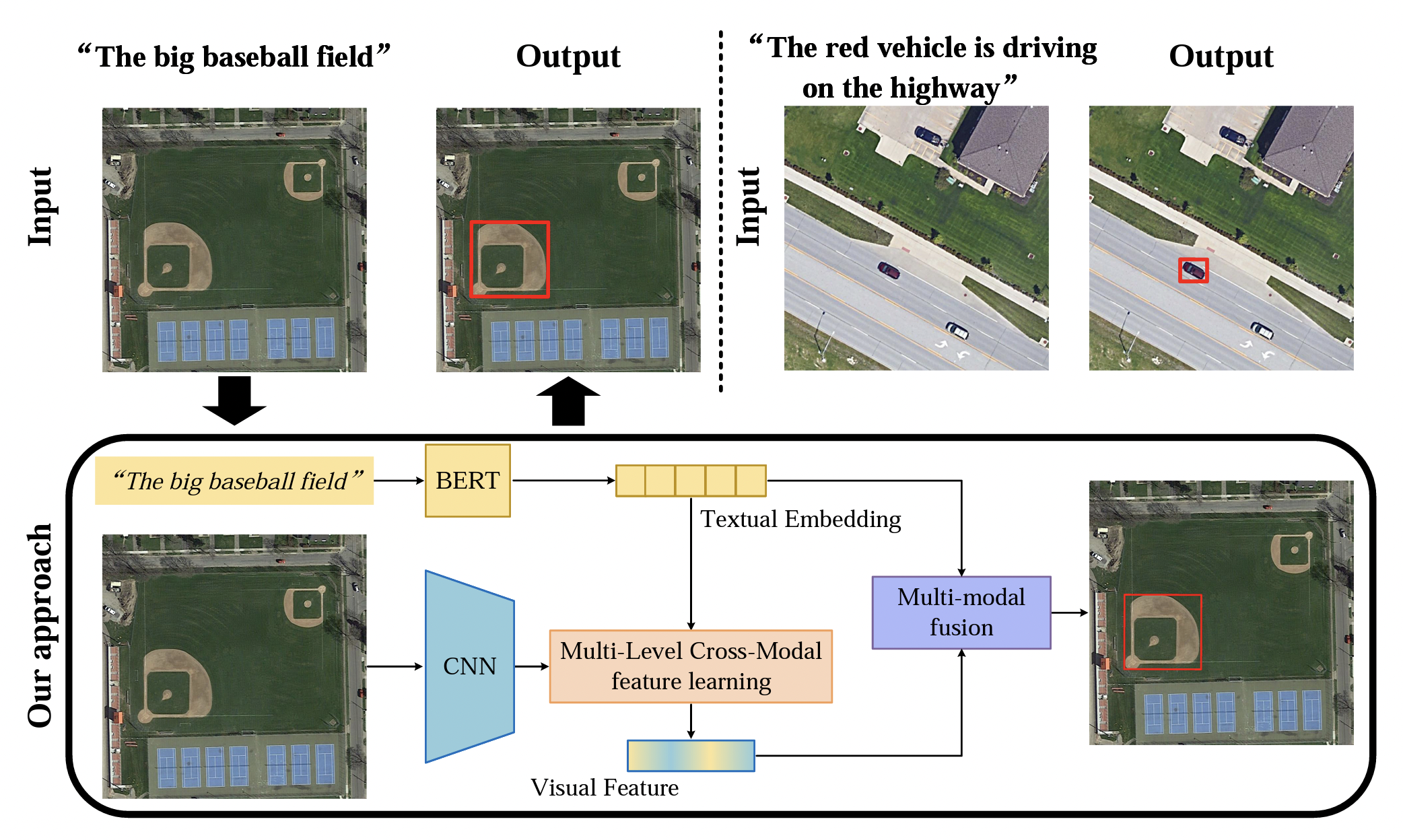}
     \caption{Illustration of remote sensing visual grounding method proposed in~\cite{zhan2023rsvg}.}
     \label{fig_visual_grounding}
\end{figure}
 
\subsection{Visual Grounding}
Visual grounding in the context of remote sensing data (RSVG) has emerged as a novel area of investigation in recent times. Studies pertaining to this specific task are currently in a nascent stage and remain relatively scarce. Specifically, RSVG involves utilizing a remote sensing image and an associated query expression to provide the bounding box for the particular object of interest~\cite{zhan2023rsvg}. Through the process of localizing objects in remote sensing scenes using natural language guidance, RSVG offers object-level understanding and facilitates accessibility for end users. The potential applications of RSVG include target object detection and recognition, search and rescue tasks, urban planning, and so on. 

Compared with query expressions in natural images, expressions in RSVG often involve complicated geospatial relations, and the objects of interest are often not visually salient. Despite the extensive research on visual grounding in natural images, this task remains under-explored in the field of remote sensing. RSVG was first introduced in~\cite{sun2022visual}, which not only introduced a novel dataset but also proposed a new method for achieving visual grounding in remote sensing scenes. The RSVG dataset contains 4,239 Google Earth images and 7,993 referring expressions with an average expression length of 28.41. The model proposed in this work consists of three components, namely, the language encoder, image encoder, and fusion module. The language encoder is used to create a geospatial relation graph, where nodes represent ground objects and edges characterize their geospatial relationships. As for the image encoder, an adaptive region attention module is leveraged to extract visual features from a large-scale remote sensing scene. In the end, a fusion module is utilized to integrate the geospatial relation graph into visual features. The proposed method achieved a grounded localization accuracy of 59.40\% and 64.95\% at the IoU threshold of 0.5 and 0.25 respectively. Another recent work was proposed by Zhan et al.~\cite{zhan2023rsvg}, in which they built a new dataset called DIOR-RSVG and developed a novel transformer-based method, shown in Fig.~\ref{fig_visual_grounding}. The DIOR-RSVG dataset contains 17,402 RS images and 38,320 referring expressions across 20 object categories, with an average expression length of 7.47. The proposed method addresses the scale variation problem by utilizing multi-scale visual features and multi-granularity language features to learn discriminative representations. To deal with cluttered backgrounds, they dynamically filter out irrelevant noise and strengthen salient features. Besides, this work also benchmarked the performance of several state-of-the-art visual grounding methods designed for natural images on remote sensing data. The proposed MGVML method achieved the best mIoU of 68.04\% on the test set of the DIOR-RSVG dataset. 

Table~\ref{table_vg} summarizes the performance of comparing methods on this dataset. We report precision scores at different thresholds and mean IoU and cumulative IoU in the Table. Please refer to ~\cite{zhan2023rsvg} for these evaluation metrics. Despite the promising capabilities of VLMs for remote sensing visual grounding, current research indicates that there is still room for improvement in the performance of RSVG methods compared to conventional object detection methods. This observation highlights the need for further research efforts to enhance the effectiveness and accuracy of RSVG methods.

\begin{table*}[htbp] 
\centering
\begin{tabular}{lccccccccc}
\toprule
Methods & Visual Encoder & Language Encoder & Pr@0.5 & Pr@0.6 & Pr@0.7 & Pr@0.8 & Pr@0.9 & meanIoU & cumIoU \\ 
\midrule
\multicolumn{2}{l}{\textbf{One-stage:}} & & & & & & & \\
ZSGNet~\cite{sadhu2019zero} & VGG & BiLSTM & 48.12 & 43.79 & 36.82 & 25.04 & 6.62 & 40.11 & 46.11 \\
ZSGNet~\cite{sadhu2019zero} & ResNet-50 & BiLSTM & 51.67 & 48.13 & 42.30 & 30.11 & 10.15 & 44.12 & 51.15 \\
FAOA-No Spatial~\cite{yang2019fast} & DarkNet-53 & BERT & 63.63 & 61.20 & 56.92 & 50.15 & 38.83 & 57.53 & 63.14 \\
FAOA~\cite{yang2019fast} & DarkNet-53 & BERT & 67.21 & 64.18 & 59.23 & 50.87 & 34.44 & 59.76 & 64.13 \\
FAOA~\cite{yang2019fast} & DarkNet-53 & LSTM & 70.86 & 67.37 & 62.04 & 53.19 & 36.44 & 60.74 & 66.20 \\
ReSC~\cite{yang2020improving} & DarkNet-53 & BERT & 72.71 & 68.92 & 63.01 & 53.70 & 33.37 & 64.24 & 68.10 \\
LBYL-Net~\cite{huang2021look} & DarkNet-53 & BERT & 73.78 & 69.22 & 65.56 & 47.89 & 15.69 & 65.92 & 73.67 \\
\midrule
\multicolumn{2}{l}{\textbf{Transformer-based:}} & & & & & & & \\
TransVG~\cite{deng2021transvg} & ResNet-50 & BERT & 72.41 & 67.38 & 60.05 & 49.10 & 20.91 & 58.57 & 76.27 \\
VLTG~\cite{yang2022improving} & ResNet-50 & BERT & 69.41 & 65.16 & 58.44 & 45.56 & 24.17 & 57.27 & 71.78 \\
VLTG~\cite{yang2022improving} & ResNet-101 & BERT & 75.79 & 72.22 & 66.33 & 56.17 & 33.31 & \textbf{69.11} & 77.85 \\
MGVLF~\cite{zhan2023rsvg} & ResNet-50 & BERT & \textbf{76.78} & \textbf{72.68} & \textbf{66.74} & \textbf{56.42} & \textbf{35.07} & 68.04 & \textbf{78.41} \\
\bottomrule
\end{tabular}
\caption{Remote sensing visual grounding performance on the DIOR-RSVG dataset. Numbers are borrowed from~\cite{zhan2023rsvg}.}
\label{table_vg}
\end{table*}

\subsection{Zero-Shot Scene Classification.}

\begin{figure*}[ht]
    \centering
    \includegraphics[width=16cm]{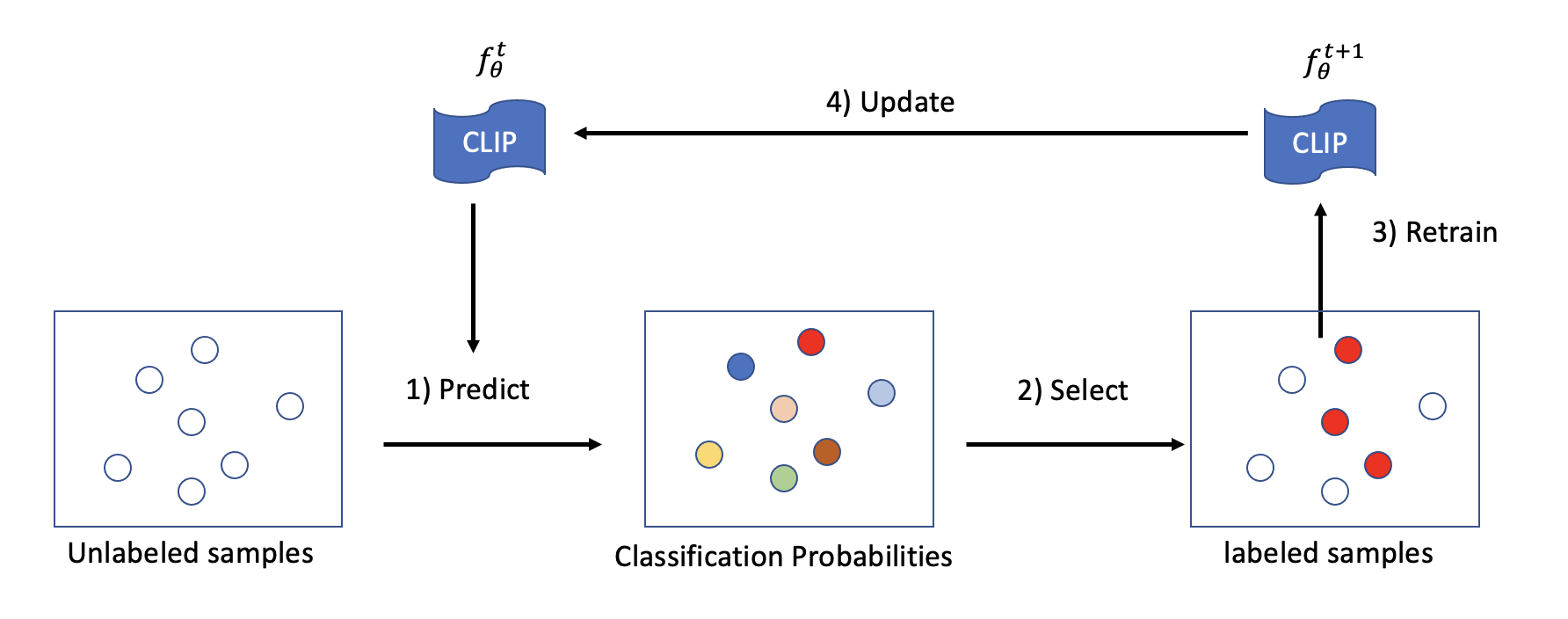}
    \caption{Overview of zero-shot remote sensing scene classification method proposed in~\cite{li2023rs}.}
    \label{fig_zero_shot_cls}
\end{figure*}

Zero-shot remote sensing scene classification (RSSC) aims to recognize unseen scene concepts by referring to both visual features and semantic relationships between semantic classes. Li et al.~\cite{li2017zero} proposed the first zero-shot learning-based approach for remote sensing (RS) scene classification, which employed a pretrained word2vec~\cite{wang2016relational} model on the Wikipedia corpus to derive semantic embeddings for category names, followed by the construction of a semantic graph to capture the inter-class relationships. The proposed ZSSC method achieved an overall classification accuracy of 49.01\% and 47.00\% on the UCM21 dataset (5 unseen classes) and the NWPU45 dataset (10 unseen classes). Quan et al.~\cite{quan2018structural} enhanced this method by incorporating a semi-supervised Sammon embedding algorithm~\cite{sammon1969nonlinear} to align the semantic and visual prototypes. The proposed ZSSC method achieved an overall classification accuracy of 50.42\% and 48.40\% on the UCM21 and NWPU45 datasets. Additionally, Sumbul et al.~\cite{sumbul2017fine} presented a zero-shot learning technique for fine-grained RS image classification, which trained a compatibility function to establish the association between image features and semantic embeddings, enabling knowledge transfer from seen to unseen classes. Building upon the work of Kodirov et al.~\cite{kodirov2017semantic}, Wang et al.~\cite{wang2021distance} proposed a distance-constrained semantic autoencoder to align the visual features and semantic representations for the zero-shot RSSC. In~\cite{li2021learning}, Li et al. employed transformer-based large language models, such as BERT, to extract semantic embeddings from expert-defined text descriptions for each class. 

Instead of using discriminative models for RSSC, Li et al.~\cite{li2022generative} introduced a generative adversarial network (GAN)-based approach for zero-shot RSSC, where a generator network was trained to synthesize image features from class semantics. The proposed method achieved a zero-shot classification accuracy of 62.66\%, 55.86\%, and 50.66\% on UCM21, AID30, and NWPU45 datasets respectively. In addition, the authors investigated different language processing models, i.e., Word2vec~\cite{wang2016relational}, Fasttest~\cite{joulin2016bag,bojanowski2017enriching}, Glove~\cite{xian2016latent}, and BERT~\cite{devlin2018bert}, for semantic embedding extraction. In contrast to natural images, RS scenes exhibit substantial structural and background variations, posing a significant challenge for models to learn robust visual features for scene understanding. Furthermore, the semantic correlations among diverse scene classes in RS images are comparatively weaker than those in natural images, rendering the semantic-reasoning-based zero-shot learning task even more challenging. 

Inspired by the great success of large vision-language models for (few-/zero-shot) natural image understanding, Li et al. introduced a large vision-language model, named RS-CLIP~\cite{li2023rs}, for remote sensing scene classification. The proposed method leveraged the pretrained CLIP model to generate pseudo labels and developed a curriculum learning strategy to improve the scene classification performance progressively. Fig. \ref{fig_zero_shot_cls} shows the method overview of the proposed method for zero-shot RSSC. Table~\ref{tab_sc} summarizes the performance of prevalent remote sensing scene classification methods. We report overall accuracy on novel classes in the Table.

\begin{table}[!htbp]
\small
\centering
\scalebox{0.95}{
\begin{tabular}{lccc}
\toprule
\multirow{2}{*}{\textbf{Method}} & UCM & NWPU & AID \\
& (5/21) & (10/45) & (5/30) \\
\midrule
% Fu et al. \cite{fu2015zero} & 47.2 & - & - \\
SAE \cite{kodirov2017semantic} & 49.50 & 44.81 & 47.34 \\
ZSP-LP \cite{li2017zero} & 49.01 & 47.00 & 46.77 \\
% ZSSC \cite{li2017zero}    & 58.7 & - & - \\
ZSC-SA \cite{quan2018structural} & 50.42 & 48.40 & 50.87 \\
WDVSc \cite{wan2019transductive} & 55.91 & 50.68 & 52.61 \\
RBGN \cite{xing2020robust} & 57.93 & 44.60 & 51.99 \\
DASE \cite{wang2021distance} & 58.63 & 51.52 & 53.49 \\
CSPWGAN \cite{li2022generative} & 62.66 & 51.52 & 55.86 \\
RS-CLIP~\cite{li2023rs} & \textbf{95.54} & \textbf{96.95} & \textbf{93.34} \\
\bottomrule
\end{tabular}
}
\caption{Zero-shot remote sensing scene classification results on four benchmark datasets. Numbers in the bracket show the number of novel classes and all classes. All values are borrowed from~\cite{li2022generative} and~\cite{li2023rs}. Boldface values indicate the best performance.}
\label{tab_sc}
\end{table}

\subsection{Few-Shot Object Detection.} 
Object detection is a prominent task in remote sensing (RS) that involves detecting object instances by identifying their bounding boxes and class labels. This field has undergone significant advancements due to extensive research efforts in recent years, including two-stage detectors, such as Fast R-CNN~\cite{fast-rcnn} and Faster R-CNN~\cite{faster-rcnn}, and one-stage detectors, such as SSD~\cite{ssd}, YOLO~\cite{yolo}, and RetinaNet~\cite{retinanet}, and recently proposed DETR variants~\cite{detr,deformable-detr}. Few-shot object detection (FSOD) in RS images is a relatively new research area that has gained significant attention in recent years. This task aims to detect objects of interest in RS images using only a few annotated examples. The pioneering work by Li et al.~\cite{li2021few} represents the initial foray into the realm of few-shot object detection specifically tailored for remote sensing images. Existing FSOD methods can be roughly divided into two families: 1) meta-learning-based~\cite{li2021few,cheng2021prototype,zhou2022fsods} and finetuning-based~\cite{zhao2021few,huang2021few,zhang2023generalized,li2023few}. Meta-learning-based methods contain a meta-learner to learn task-agnostic knowledge from a large number of sampled tasks, each task (also called episode) includes a support set and a query set and uses a task-specific learner to perform detection on specific tasks. Finetuning-based methods first train a detection model on base classes and then finetune network weights on novel classes. A comprehensive literature review of few-shot object detection in RS images can be found in~\cite{liu2022few}. 

\begin{figure*}[ht]
    \centering
    \includegraphics[width=0.9\linewidth]{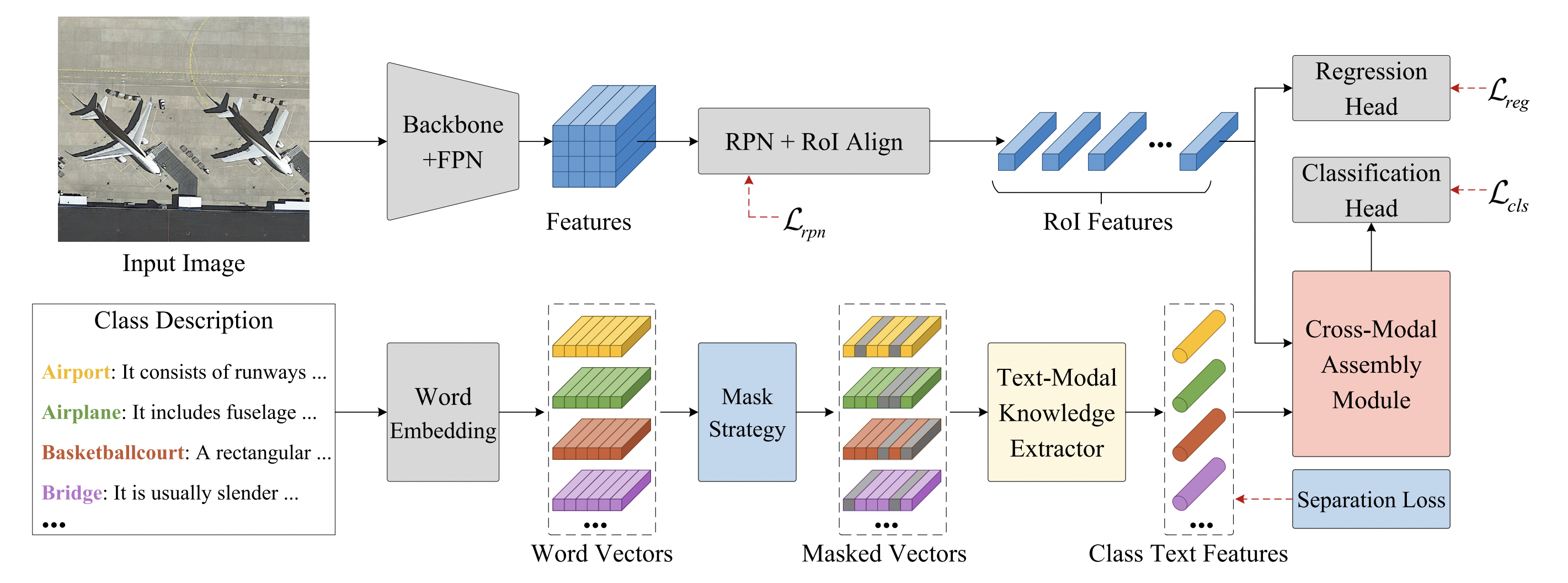}
    \caption{The architecture of the few-shot object detection method proposed in~\cite{lu2023few}. Class descriptions are used for object classification. }
    \label{fig_few_shot_det}
\end{figure*}

Only a few attempts exploit VLMs for FSOD in RS images. Zhang et al.~\cite{zhang2023text} proposed to build a corpus that contains language descriptions for each region, such as object attributes and relations, to encode the corresponding common sense embeddings. The proposed TSF-RGR method achieved a mAp of 0.77 and 0.49 on 10 novel classes of NWPU VHR-10 and DIOR datasets respectively. Their results also show that text-based semantic relations can significantly improve the few-shot detection performance. Lu et al.~\cite{lu2023few} proposed to use text descriptions for all object categories as additional features to mitigate classification confusion in FSOD. Fig. \ref{fig_few_shot_det} shows the method overview of the proposed method for few-shot object detection in RS images. The proposed TEMO method achieved a mAp of 0.75 and 0.43 on 10 novel classes of NWPU VHR-10 and DIOR datasets respectively. These methods show the potential of combining visual and language features for better representation learning and thus enhance the performance of FSOD in RS images. Table~\ref{table_od} summarizes the performance of prevalent few-shot object detection methods for remote sensing images. We report the average precision (AP) of each novel class and the mean AP in the table. PL, BD, and TC represent plane, baseball diamond, and tennis court, respectively.

\begin{table*} 
\centering
\begin{tabular}{llccccccccccccc}
\toprule
\multirow{2}{*}{Shot} & \multirow{2}{*}{Methods} & \multicolumn{6}{c}{DIOR dataset} & & \multicolumn{4}{c}{NWPU VHR-10 dataset} \\
\cline{3-8}\cline{10-13}
& & airplane & baseball field & tennis court & train station & windmill & mAP & &  PL & BD & TC & mAP \\
\midrule
% \multirow{5}{*}{2} & Meta RCNN$^\dagger$ \cite{yan2019meta} & 9.1 & 15.4 & 33.6 & 10.1 & 4.3 & 14.5 \\
% & FsDetView$^\dagger$ \cite{xiao2022few} & 9.1 & 9.1 & 18.0 & 9.1 & 0.1 & 9.1 \\
% & TFA$^\dagger$ \cite{wang2020frustratingly} & 12.0 & 24.6 & 33.7 & 8.9 & 12.2 & 18.3 \\
% & FSCE$^\dagger$ \cite{sun2021fsce} & 20.7 & 38.5 & 42.2 & 10.0 & 14.0 & 25.1 \\
% & TEMO$^\dagger$ \cite{lu2023few} & \textbf{24.4} & \textbf{43.0} & \textbf{43.1} & \textbf{11.4} & \textbf{15.4} & \textbf{27.5} \\
% \hline
\multirow{5}{*}{3} & Meta RCNN$^\dagger$ \cite{yan2019meta} & 9.1 & 17.8 & 33.5 & 3.8 & 9.1 & 14.7 & & 6.8 & 7.1 & 5.7 & 6.5\\
& FsDetView$^\dagger$ \cite{xiao2022few} & 9.1 & 17.9 & 18.1 & 9.1 & 9.1 & 12.6 & & 9.1 & 22.9 & 2.4 & 11.5\\
& TFA$^\dagger$ \cite{wang2020frustratingly} & 11.9 & 22.6 & \textbf{51.8} & 9.7 & 8.1 & 20.8 & & 9.1 & 16.2 & 14.5 & 13.2\\
& Meta YOLO \cite{kang2019few} & - & - & - & - & - & - & & 13.0 & 12.0 & 11.0 & 12.0\\
& FSODM \cite{li2021few} & - & - & - & - & - & - & & 15.0 & 57.0 & 25.0 & 32.0 \\
& PAMS-Det \cite{zhao2021few} & 14.0 & \textbf{54.0} & 24.0 & 17.0 & \textbf{31.0} & 28.0 & &21.0 & 76.0 & 16.0 & 37.0 \\
& FSCE$^\dagger$ \cite{sun2021fsce} & 17.3 & 38.7 & 40.8 & 11.6 & 16.0 & 24.9 
 & & 33.3 & 73.4 & 30.2 & 45.6 \\
& DMLKI \cite{li2023few} & - & - & - & - & - & - & & 41.0 & 66.0 & 29.0 & 45.0 \\
& TEMO$^\dagger$ \cite{lu2023few} & \textbf{18.4} & 47.1 & 52.4 & \textbf{17.7} & 18.8 & \textbf{30.9} & & 49.1 & 77.8 & \textbf{45.8} & \textbf{57.6}\\
& TSF-RGR \cite{zhang2023text} & - & - & - & - & - & - & & \textbf{57.0} & \textbf{79.0} & 35.0 & 57.0 \\
\midrule
\multirow{11}{*}{5} & Meta RCNN$^\dagger$ \cite{yan2019meta} & 9.1 & 25.9 & 32.7 & 7.0 & 2.3 & 15.4 & &12.0 & 21.9 & 0.9 & 11.6\\
& FsDetView$^\dagger$ \cite{xiao2022few} & 9.1 & 22.3 & 26.3 & 5.0 & 1.1 & 12.8 & &16.5 & 28.4 & 7.8 & 17.6 \\
& TFA$^\dagger$ \cite{wang2020frustratingly} & 13.2 & 27.7 & 51.8 & 9.0 & 8.0 & 21.9  & &9.1 & 15.9 & 24.6 & 16.5\\
& Meta YOLO \cite{kang2019few} & 9.0 & 33.0 & 47.0 & 9.0 & 13.0 & 22.0 & & 24.0 & 39.0 & 11.0 & 24.0\\
& RepMet \cite{karlinsky2019repmet} & 9.0 & 19.0 & 11.0 & 1.0 & 1.0 & 8.0 & & 20.0 & 36.0 & 14.0 & 23.0 \\
& FSODM~\cite{li2021few} & 9.0 & 27.0 & 57.0 & 11.0 & 19.0 & 25.0  & &58.0 & 84.0 & 16.0 & 53.0\\
& PAMS-Det \cite{zhao2021few} & 17.0 & 55.0 & 41.0 & 17.0 & \textbf{34.0} & 33.0 & &55.0 & \textbf{88.0 }& 20.0 & 55.0 \\
& FSCE$^\dagger$ \cite{sun2021fsce} & 22.8 & 36.6 & 49.5 & 17.1 & 15.2 & 28.3 & &60.7 & 83.6 & 32.5 & 59.0 \\
& DMLKI \cite{li2023few} & 15.0 & 46.0 & 48.0 & 17.0 & 17.0 & 28.0 &  & 53.0 & 81.0 & \textbf{49.0} & 61.0 \\
& TEMO$^\dagger$ \cite{lu2023few} & 24.2 & 58.9 & 52.1 & \textbf{25.0} & 25.7 & 37.2  & & \textbf{71.4} & 85.4 & 43.5 & \textbf{66.8}\\
& TSF-RGR \cite{zhang2023text} & \textbf{58.0} & \textbf{79.0} & \textbf{58.0} & 11.0 & 6.0 & \textbf{42.0} & & 71.0 & 85.0 & 42.0 & 66.0 \\
\midrule
\multirow{10}{*}{10} & Meta RCNN$^\dagger$ \cite{yan2019meta} & 12.4 & 37.5 & 32.4 & 14.0 & 3.2 & 19.9 && 11.4 & 24.4 & 10.5 & 15.4\\
& FsDetView$^\dagger$ \cite{xiao2022few} & 9.1 & 31.7 & 25.8 & 11.1 & 6.1 & 16.7  & & 9.1 & 20.0 & 12.4 & 13.8\\
& TFA$^\dagger$ \cite{wang2020frustratingly} & 17.7 & 33.1 & 56.4 & 17.3 & 8.7 & 26.7 & & 4.5 & 17.0 & 21.9 & 14.5 \\
& Meta YOLO \cite{kang2019few} & 15.0 & 45.0 & 54.0 & 7.0 & 18.0 & 28.0 &  & 20.0 & 74.0 & 26.0 & 40.0 \\
& RepMet \cite{karlinsky2019repmet} & 13.0 & 33.0 & 24.0 & 1.0 & 1.0 & 14.0 &  & 22.0 & 39.0 & 18.0 & 26.0\\
& FSODM~\cite{li2021few} & 16.0 & 46.0 & 60.0 & 14.0 & 24.0 & 32.0  & & 60.0 & 88.0 & 48.0 & 65.0\\
& PAMS-Det \cite{zhao2021few} & 25.0 & 58.0 & 50.0 & 23.0 & \textbf{36.0} & 38.0 & & 61.0 & 88.0 & 50.0 & 66.0\\
& FSCE$^\dagger$ \cite{sun2021fsce} & 24.7 & 56.2 & 49.1 & 21.6 & 31.2 & 34.8 & & 85.2 & 75.3 & 48.0 & 69.5 \\
& DMLKI \cite{li2023few} & 32.0 & 53.0 & 48.0 & 16.0 & 20.0 & 34.0 & & 77.0 & 90.0 & 68.0 & 78.0 \\
& TEMO$^\dagger$ \cite{lu2023few} & 32.6 & 59.3 & 67.0 & \textbf{23.9} & 31.2 & 42.8  & & \textbf{86.4} & 88.7 & 50.1 & 75.1\\
& TSF-RGR \cite{zhang2023text} & \textbf{72.0} & \textbf{79.0} & \textbf{67.0} & 8.0 & 19.0 & \textbf{49.0} & & 78.0 & \textbf{94.0} & \textbf{60.0} & \textbf{77.0} \\
\bottomrule
\end{tabular}
\caption{Few-shot detection performance on the novel classes of the DIOR dataset and NWPU VHR-10 dataset. $^\dagger$ denotes the average results over 20 random experiments. The boldface indicates the best performance in each column. Numbers are borrowed from~\cite{lu2023few}.}
\label{table_od}
\end{table*}

In computer vision, VLMs have shown great potential in zero-shot object detection. The utilization of multi-modal vision-language pre-training models has made Open-Vocabulary Object Detection (OVOD) an active research area, as it allows for more realistic scenarios to be considered. In contrast to traditional object detection methods that are trained and evaluated on fixed and predefined classes, OVOD involves training on annotated datasets and generalizing the trained models to previously unseen novel classes. To enable open-vocabulary object detection, a common strategy is to modify existing object detection heads by matching object features and class text embeddings~\cite{radford2021learning,guzero,zheng2020background,rahman2020improved,bansal2018zero}. In remote sensing, this open-vocabulary setting opens up a new door for researchers to develop VLMs to detect objects from unseen categories in a more generalizable and practical setting. 

\begin{figure*}
    \centering
    \includegraphics[width=16cm]{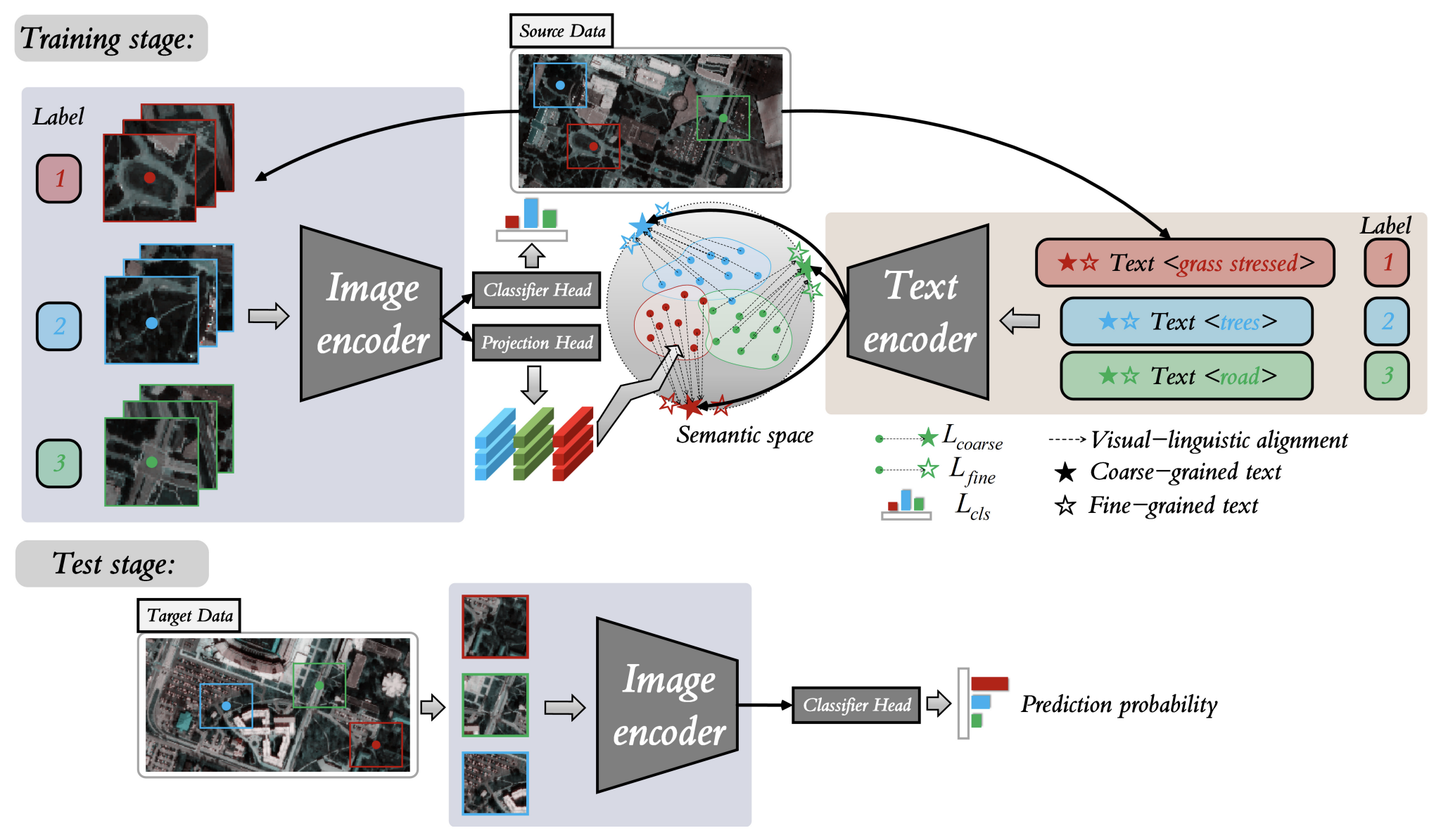}
    \caption{Overview of cross-scene hyperspectral image classification method proposed in~\cite{zhang2023language}. Visual features and text features are aligned in the semantic space.}
    \label{fig_sem_seg}
\end{figure*}

\subsection{Few-/Zero-shot Semantic Segmentation}
In the realm of semantic segmentation, few-shot learning approaches enable the segmentation of novel classes with a limited number of annotated images. Recent efforts have focused on two categories, namely parameter matching-based and prototype-based methods. Notably, the Pioneering work PANet~\cite{wang2019panet} achieved a breakthrough in few-shot segmentation by introducing a prototype alignment module that produces highly representative prototypes for each semantic class and segments query objects based on feature matching. Jiang et al.~\cite{jiang2022few} proposed the first few-shot learning method for remote sensing image segmentation and showed the possibility of conducting RSI semantic segmentation using only a few samples from novel classes. The proposed method achieved an overall classification accuracy of 67.7\% and 53.2\% on ISPRS Vaihingen and Potsdam datasets~\cite{isprs_dataset} with only 5 annotated patches.

To overcome the data reliance on deep learning-based segmentation methods, recent studies~\cite{chen2022semi,zhang2023language} have explored self-/semi-supervised learning and weakly supervised learning to reduce the need for dense annotation. Chen et al.~\cite{chen2022semi} introduced a semi-supervised method for the few-shot segmentation of RS images based on contrastive learning. Zhang et al. \cite{zhang2023language} introduced a network for cross-scene hyperspectral image classification, which utilizes language guidance to achieve domain generalization. Fig. \ref{fig_sem_seg} gives an overview of the proposed method. 
Lang et al.~\cite{lang2023global} introduced a remote sensing few-shot segmentation framework based on global rectification and decoupled registration and demonstrated superior performance than previous SOTA methods on the iSAID-5i dataset~\cite{waqas2019isaid}. Although these methods have shown the potential of applying VLMs for the few-shot semantic segmentation of RSIs, the adoption of VLMs in RS image segmentation remains in its infancy. Table~\ref{table_sem_seg} summarizes the performance of few-shot semantic segmentation methods for remote sensing images. We report class-average IoU in the Table.

In computer vision, pre-trained VLMs have been extensively explored for open-vocabulary semantic segmentation, thanks to their remarkable success in open-vocabulary image classification, in which the model can classify any category without the need for additional annotated images of that category.
Inspired by the great success of CLIP, DenseCLIP~\cite{denseclip} solved the dense prediction problem by matching each pixel with the text-based class descriptions.
MaskCLIP~\cite{maskclip} employed pseudo per-pixel labels created from CLIP and self-training to achieve annotation-free segmentation.
Similarly,~\cite{zabari2021semantic} used pixel-level pseudo-labels for dense supervision. 
CLIPSeg~\cite{clipseg} built upon the CLIP model and produced image segmentations based on arbitrary prompts.
ZSSeg~\cite{zsseg} first generated mask proposals and then utilized CLIP to classify the generated proposals in a two-stage manner.
LSeg~\cite{lseg} utilized an image encoder to match pixel embeddings with text embeddings and a text encoder to offer a flexible class representation.
OVSeg~\cite{ovseg} proposed to finetune the CLIP model on masked image regions and corresponding texts from the COCO Captions dataset and obtained better open vocabulary segmentation performance.
Fusioner~\cite{fusioner} combined different vision and language models using a cross-modality fusion module to achieve open-vocabulary semantic segmentation.
SegCLIP~\cite{segclip} gathered patches with learnable centers to semantic regions through text-image pair training, which can dynamically capture semantic groups and generate segmentation results.
Given the prosperity of open-vocabulary semantic segmentation for natural images, we believe this can be a promising research direction in the remote sensing community. 

\begin{table*}[htbp]
\centering
\begin{tabular}{lcccccccc}
\toprule
\multirow{2}{*}{Method} & \multirow{2}{*}{Year} & \multicolumn{3}{c}{1-shot} & & \multicolumn{3}{c}{5-shot} \\
\cline{3-5}\cline{7-9}
& & Fold-0 & Fold-1 & Fold-2 & & Fold-0 & Fold-1 & Fold-2 \\
\midrule
PANet \cite{wang2019panet} & 2019 & 27.56 & 17.23 & 24.60 & &  36.54 & 16.05 & 26.22 \\
CANet~\cite{zhang2019canet} & 2019 & 25.51 & 13.50 & 24.45 & & 29.32 & 21.85 & 26.03 \\
SCL \cite{zhang2021self} & 2021 & 34.78 & 22.77 & 31.20 & & 41.29 & 25.23 & 31.91 \\
PFENet~\cite{tian2020prior} & 2022 & 35.84 & 23.35 & 27.20 & & 44.42 & 25.34 & 33.59 \\
NERTNet~\cite{liu2022learning} & 2022 & 34.93 & \textbf{23.95} & 28.56 & & \textbf{48.38} & 26.73 & 37.19 \\
DCP \cite{lang2022beyond} & 2022 & 37.83 & 22.86 & 28.92 & & 41.52 & \textbf{28.18} & 33.43 \\
BAM~\cite{lang2022learning} & 2022 & 39.43 & 21.69 & 28.64 & & 29.32 & 27.92 & 38.62 \\
DMML \cite{wang2021dmml} & 2022 & 28.45 & 21.02 & 23.46 & & 30.61 & 23.85 & 24.08 \\
SDM~\cite{yao2021scale} & 2022 & 27.96 & 21.99 & 27.82 & & 28.50 & 25.23 & 31.07 \\
DML \cite{jiang2022few} & 2022 & 32.96 & 18.98 & 26.27 & & 33.58 & 22.05 & 28.77 \\
TBPNet \cite{puthumanaillam2023texture} & 2022 & 29.33 & 16.84 & 25.47 & & 30.38 & 20.42 & 29.07 \\
R\textsuperscript{2}Net \cite{lang2023global} & 2023 & \textbf{41.22} & 21.64 & \textbf{35.28} & & 46.45 & 25.80 & \textbf{39.84} \\
\bottomrule
\end{tabular}
\caption{Few-shot semantic segmentation performance on the iSAID-5i dataset in mIoU under 1-Shot and 5-Shot Settings. Numbers are borrowed from~\cite{lang2023global}. All comparing methods use a ResNet-50 Backbone.}
\label{table_sem_seg}
\end{table*}

\section{Useful Resources}
In this section, we summarize the commonly used remote sensing vision language datasets, codebases, and other resources.

\subsection{RS vision-language datasets}
We summarize remote sensing vision language datasets in Table \ref{tab_dataset}. Although existing works mostly use scene classification datasets for building remote sensing foundation models, all listed datasets can potentially be used for building general-proposal remote sensing foundation models. Note that we only list the commonly used datasets and may not fully cover all related datasets for each task. A comprehensive review of remote sensing datasets can be found at~\cite{xiong2022earthnets}.

\begin{table*}[htp]
    \centering
    % \resizebox{0.9\textwidth}{!}{%
    \begin{tabular}{p{2.5cm}p{12cm}p{2.5cm}}
        \toprule
        Dataset & Description & Tasks\\
        \midrule
        UCM \cite{yang2010bag} & {The UC Merced land use dataset contains 2,100 aerial images from 21 scene categories, and each category contains 100 images of size $256 \times 256$.} & Scene Classification, Foundation Model\\
        Sydney dataset~\cite{zhang2014saliency} & 613 high-resolution remote sensing images from 7 different scene categories.  & Scene Classification, Foundation Model \\
        WHU-RS19 \cite{Dai2011WHURS19} & The WHU-RS19 dataset contains aerial 1,013 images from 19 scene categories. There are, in total, 1,013 images with $600 \times 600$ pixels.  & Scene Classification, Foundation Model \\
        NWPU-RESISC45 \cite{cheng2017remote} & The NWPU-RESISC45 dataset consists of 31,500 aerial images from 45 scene categories, and each category has 700 images with a size of $256 \times 256$ pixels.  & Scene Classification, Foundation Model  \\
        AID \cite{xia2017aid} & The AID dataset contains 10,000 images from 30 scene categories, and each image is of size $600 \times 600$ pixels. & Scene Classification, Foundation Model \\
        SEN12MS \cite{Schmitt2019} & SEN12MS is a large-scale multi-sensor remote sensing dataset with 180,662 triplets of dual-pol synthetic aperture radar (SAR) image patches, multi-spectral Sentinel-2 image patches, and MODIS land cover maps. & Scene Classification, Foundation Model \\
        BigEarthNet \cite{sumbul2021bigearthnet} & BigEarthNet is a multi-modal multi-label remote sensing imagery dataset that contains 590,326 pairs of Sentinel-1 and Sentinel-2 image patches. & Scene Classification, Image Retrieval, Foundation Model.
        \\
        MillionAID \cite{million-aid} & A large-scale remote sensing imagery dataset with a million images from 51 semantic scene categories. & Scene Classification, Foundation Model \\
        SatlasPretrain \cite{bastani2023satlaspretrain} & It integrates Sentinel-2 and NAIP imagery with 302M annotations across 137 categories and 7 types of labels. & Scene Classification, Foundation Model \\
        \midrule
        UCM-Captions~\cite{qu2016deep} & 2,100 image-caption pairs based on the UCM dataset. Each image is described using five distinct sentences. & Image Caption, Text-to-Image Retrieval. \\
        Sydney-Captions~\cite{qu2016deep} & 613 image-caption pairs based on the Sydney dataset. Each image is described using five distinct sentences.  & Image Caption, Text-to-Image Retrieval \\
        RSICD \cite{lu2017exploring} & The RSICD dataset comprises 10,921 images sourced from platforms like Google Earth, Baidu Map, MapABC, and Tianditu. Each image, standardized to a resolution of $224 \times 224$ pixels, is accompanied by five descriptive sentences.  & Image Caption, Text-to-Image Generation / Retrieval \\
        RS5M \cite{zhang2023rs5m} & This dataset contains 5M remote sensing images with machine-generated captions. The dataset is built by filtering publicly available image-text paired datasets and exclusively captioning RS datasets using a pre-trained VLM.  & Image Caption \\
        RSICap \cite{hu2023rsgpt} & The RSICap dataset comprises 2,585 high-quality human-annotated captions with rich information.  & Image Caption \\
        TextRS \cite{abdullah2020textrs} & {The TextRS dataset compiles 2,144 images from four distinct scene datasets: AID, Merced, PatternNet, and NWPU datasets. Each image comes with descriptive sentences provided by five different annotators.} & Text-to-Image Retrieval \\
        \midrule
        RSVQA \cite{lobry2020rsvqa} & RSVQA-HR contains 10,569 high-resolution images and 1,066,316 question-answer pairs; RSVQA-LR contains 772 low-resolution images and 77,232 question-answer pairs.  & Visual Question Answering\\
        RSIVQA \cite{zheng2021mutual} & RSIVQA utilizes two types of annotation methods: manual annotation and automatic generation, resulting in a dataset comprising over 110,000 VQA triplets. & Visual Question Answering\\
        CDVQA \cite{yuan2022change} & CDVQA is a change detection visual question answering dataset, which contains 2968 pairs of aerial images and more than 122,000 corresponding question-answer pairs. & Visual Question Answering\\
        VQA-TextRS \cite{al2022open} & {VQA-TextRS is a dataset consisting of 2,144 RS images from four source datasets, with each image having 2-3 English question/answer pairs, totaling 6,245 pairs, categorized into four question classes.} & Visual Question Answering\\
        RSIEval \cite{hu2023rsgpt} & RSIEval consists of 100 high-quality image-caption pairs with one caption per image, and 936 diverse image-question-answer triplets with an average of 9 questions per image.  & Visual Question Answering \\
        \midrule
        RSVG~\cite{sun2022visual} & The dataset contains 4,239 remote sensing images with 7,933 textual annotations for 5,994 objects. Each image has an average of 1.87 referring expressions with an average length of 28.41 words. & Visual Grounding \\
        DIOR-RSVG \cite{zhan2023rsvg} & The DIOR-RSVG dataset encompasses 38,320 linguistic descriptions spanning 17,402 remote sensing images from the DIOR dataset, covering 20 distinct object categories. Expressions have an average length of 7.47 words, with a vocabulary set comprising 100 terms. & Visual Grounding\\
        \midrule
        NWPU VHR-10 \cite{cheng2014multi} & This dataset offers 800 aerial images across 10 classes with a spatial resolution between 0.5 to 2 m. 650 of these images have annotated objects such as airplanes, ships, storage tanks, and various sports courts. & Object Detection \\
        DIOR \cite{li2020object} & This dataset contains 23,463 aerial images of 0.5 to 30 m resolution, featuring 192,472 object instances from 20 classes, such as airplanes, courts, bridges, harbors, vehicles, etc.  & Object Detection  \\
        FAIR1M \cite{sun2022fair1m} & The FAIR1M dataset comprises over 1 million instances across more than 40,000 images with spatial resolutions between 0.3m and 0.8m. Every object within this dataset is annotated, falling into 5 primary categories and further divided into 37 sub-categories. & Object Detection  \\%Image sizes in the dataset vary, ranging from 1,000 x 1,000 to 10,000 x 10,000 pixels, showcasing objects with diverse scales, orientations, and forms.  & Object Detection
        \midrule
        ISPRS Vaihingen/Potsdam~\cite{rottensteiner2012isprs} & The ISPRS Vaihingen (resp. Potsdam) dataset comprises 33 (resp. 38) ultra-high-resolution aerial images, with a ground sampling distance of 9cm (resp. 5cm). Each pixel in the image dataset is labeled into six categories: buildings, impervious surfaces, low vegetation, trees, vehicles, and clutter. & Semantic Segmentation \\
        Zurich Summer dataset\footnotemark[2] & The dataset contains 20 image crops sourced from a QuickBird capture of Zurich in August 2002. The images have been enhanced to the PAN resolution with an approximate 0.62cm GSD. Eight distinct urban and surrounding urban categories have been labeled by hand.  & Semantic Segmentation \\
        Houston~\cite{debes2014hyperspectral,le20182018} & This dataset comprises two images: a hyperspectral image and a LiDAR-based DSM, both with a 2.5m spatial resolution, captured over the University of Houston and its surrounding areas. Pixels in the image are labeled into 16 semantic categories.  & Semantic Segmentation \\
        Pavia~\cite{gamba2004collection} & Pavia Centre and Pavia University are collected based on a ROSIS sensor flight over northern Italy. The Pavia Centre image has 102 spectral bands with $1096 \times 1096$ pixels, while the Pavia University has 103 bands with $610 \times 610$ pixels. Each image is divided into 9 semantic classes.  & Semantic Segmentation \\
        GID~\cite{tong2020land} & The GID dataset contains two segments: a broad-scale categorization set and a detailed land-cover classification set. The broad-scale set includes 150 pixel-level labeled GaoFen-2 images, while the detailed set consists of 30,000 multi-scale image fragments paired with 10 pixel-level labeled GF-2 images.  & Semantic Segmentation  \\
        \bottomrule
    \end{tabular}
    % }
    \caption{Summary of commonly used datasets for remote sensing vision language tasks.}
    \label{tab_dataset}
\end{table*}

\footnotetext[2]{https://sites.google.com/site/michelevolpiresearch/data/zurich-dataset}

\subsection{Open-source Codebases}
We further list several commonly used open-source codebases to provide an easy starting point for researchers attempting to work on vision language models in remote sensing.
\begin{itemize}
\item Huggingface\footnote{https://huggingface.co/}. Huggingface is a collaborative hub for machine learning practitioners to work on models, datasets, and applications. It offers solutions to expedite machine learning projects, boasting over 50,000 organizational users. It facilitates VLM development through its Transformers library, which simplifies the process of downloading, running, and fine-tuning various vision-language models. 

\item MiniGPT-4. MiniGPT-4 \cite{zhu2023minigpt} is one of the pioneering open-source works that show extraordinary vision-language abilities, such as generating websites from handwritten text and identifying humorous elements within images. Its enhanced version, MiniGPT-v2 \cite{chen2023minigptv2} improves the model by enabling spatial understanding and visual grounding abilities. The codebase serves as a simple starter kit for researchers to develop powerful vision-language models.

\item LLaVA. Large Language and Vision Assistant (LLaVA)~\cite{liu2024visual} introduces a large multimodal model. This model merges a vision encoder with Vicuna, aiming for general-purpose visual and linguistic comprehension. It boasts remarkable chat functionality, reminiscent of the multimodal features of GPT-4, and establishes unparalleled accuracy in Science QA. Both multimodal instruction tuning data, model, and code base are publicly available. LLaVA-1.5 further improves the model and attains the best results on 11 vision-language benchmarks.

\item QWen-VL \cite{bai2023qwen}. Qwen-VL, developed by Alibaba Cloud under the Qwen series, is a multimodal model that processes image, text, and bounding box inputs to produce text and bounding box outputs. It excels over similar open-sourced large vision language models on multiple English benchmarks like Zero-shot Captioning, VQA, DocVQA, and Grounding. The model supports English, Chinese, and multilingual interactions, improving end-to-end bilingual text recognition in images.

\item Shikra \cite{chen2023shikra}. Shikra is a multimodality LLM designed to initiate referential dialogue by showcasing proficiency in handling spatial coordinate inputs/outputs through natural language, all without the need for extra vocabularies, position encoders, pre-/post-detection measures, or external plugin models.

\end{itemize}

\subsection{Other resources}
There are also many online accessible resources/tools that are useful for building vision language models in the remote sensing domain.

\begin{itemize}
    \item OpenAI API\footnote{https://platform.openai.com/}. Although OpenAI does not release its code and models for the GPT series, it provides a user-friendly API for users to call its powerful products, e.g., GPT-4. OpenAI API provides interfaces for text generation, image generation, text/image embedding, and other multi-modality tools (e.g., audio-to-text). It serves as a useful tool for researchers to prepare their own dataset~\footnote{https://sharegpt.com/}.
    \item LAVIS. LAVIS~\cite{li2023lavis} is a comprehensive Python framework tailored for integrated language and vision research. It offers a unified platform for a variety of tasks like captioning, retrieval, and visual question answering. The library encompasses over 20 standard datasets and supports over 30 pretrained VLMs. The goal is to streamline the development of multimodal models for both researchers and developers.
    \item Midjourney\footnote{https://www.midjourney.com/}. Midjourney is currently one of the most impressive tools for text-to-image generation. Midjourney significantly propels vision-language research forward by providing a powerful tool for diverse text-image pair data generation. 
    
\end{itemize}

\section{Conclusion and Future Trends}
By incorporating both visual and linguistic modalities, VLMs have the potential to revolutionize how RS data is processed and interpreted. These models empower researchers and practitioners with advanced tools to extract meaningful insights from remote sensing imagery, enabling more accurate and sophisticated analyses of Earth's surface and its dynamics. With their ability to bridge the gap between visual perception and language comprehension, VLMs offer promising avenues for the development of innovative RS applications and decision-support systems. As this field continues to evolve, the integration of VLMs into RS workflows is expected to contribute significantly to the advancement of remote sensing science and its practical applications.

In the past few years, many studies in remote sensing (RS) have demonstrated the superiority of VLMs over purely visual models in various RS tasks, including image captioning, text-based image generation, text-based image retrieval, visual question answering, scene classification, semantic segmentation, and object detection. While these early attempts have shown the success of applying VLMs to remote sensing, it is still an emerging field for most researchers. Hence, this paper presents a comprehensive review of the application of visual language models in remote sensing, providing other researchers with an in-depth understanding of the background and recent advances in this rapidly growing field. It also seeks to encourage further investigations in this exciting and promising field.

After reviewing the literature on VLMs in remote sensing, we identified several limitations in the current research. Firstly, the number of RS datasets used for training VLMs is limited, and the sample size is much smaller than the billions of image datasets in the computer vision field. Secondly, most of the existing VLMs in RS still use classical CNN and RNN as image and language encoders, with only a few works exploring pre-trained visual transformers and large language models, such as GPT, BERT, and Flan-T5. This restricts the feature learning abilities of these models. Additionally, training these VLMs from scratch requires substantial computational resources, particularly for large networks with billions of parameters. Thus there calls for effective model finetuning techniques for training large VLMs in RS. Moreover, RS data can exhibit high variability due to factors such as lighting conditions, atmospheric interference, and sensor noise. This variability makes it difficult for VLMs to capture the relationships between visual and textual information accurately. However, little existing work has taken this into account. Furthermore, VLMs may struggle to handle the large spatial and temporal scales of RS data, which can cover large areas and span long periods, making it challenging to capture the relationships between visual and textual information over space and time.

Given the limitations of existing VLMs research in RS. We list several promising research directions in the following.

\begin{itemize}
    \item \textbf{Large-scale image-text pair dataset.} It is well-known that the accuracy of AI-based systems is heavily reliant on the scale and diversity of training datasets. However, in RS, the existing largest datasets, including the Million-AID, fall short in terms of scale compared to web-scale datasets employed in the computer vision community that encompass billions of images. For instance, the LAION-5B~\cite{schuhmann2022laion} dataset is an open-source collection that comprises over 5 billion image-text pairs. To address the pressing need for more rich datasets that can facilitate the training of large VLMs in RS at scale, concerted efforts must be made to create data collection and sharing mechanisms. Therefore, it is essential for the research community to collaborate toward building datasets that are sufficiently diverse and paired with language descriptions.
    
    \item \textbf{Unified vision-language models.} Following the notable success of MiniGPT-4~\cite{zhu2023minigpt} and LLaVa~\cite{liu2024visual}, there has been a burgeoning research interest within the computer vision community in leveraging Large Language Models (LLMs) for vision-language tasks. Capitalizing on the robust reasoning capabilities of LLMs, recent multi-modal LLMs have been designed to process inputs from a variety of modalities, including images, videos, audio, and point clouds. Building on this research trend, recent initiatives such as RSGPT~\cite{hu2023rsgpt} and GeoChat~\cite{kuckreja2023geochat} have investigated the feasibility of developing unified vision-language models for a wide range of remote sensing tasks, including but not limited to image captioning, visual question answering, visual grounding, and region captioning. The exploration of more advanced, general-purpose vision-language models for remote sensing presents a promising research avenue towards enhancing the understanding of remote sensing imagery.

    \item \textbf{Text-based image generation using diffusion models.} 
    Existing neural networks usually require a considerable volume of data to be trained for convergence, however, the collection of data requires a large amount of human and material resources. Diffusion models~\cite{rombach2022high}, on the other hand, have recently attracted huge attention due to their ability to generate high-quality images with detail and high fidelity. By using diffusion models to generate new images based on existing text descriptions, we can create synthetic data and effectively expand the size of our datasets to improve the robustness and generalization of deep learning models. Additionally, by incorporating techniques such as style transfer or domain adaptation, we can generate synthetic images that are more diverse and representative of real-world scenarios, further enhancing the effectiveness of data augmentation through diffusion models.    
    
    \item \textbf{Few-/zero-shot learning.} Benefiting from the powerful reasoning abilities of LLMs, VLMs show great potential for data-efficient learning by recognizing unseen objects or patterns based on the relationships between words and concepts in vision data. This makes them particularly useful in few-/zero-shot learning scenarios, in which limited labeled data is available for training. While previous attempts have explored the understanding of RS images in few-/zero-shot settings using smaller vision and language models, they lack the reasoning abilities necessary to comprehend and identify unseen objects or patterns. As we move towards the era of AGI, new techniques must be designed to better integrate LLMs into RS image understanding tasks, such as object detection, semantic segmentation, and change detection, particularly in few-/zero-shot settings.

    \item \textbf{Efficient finetuning on RS data.} Existing large language models usually contain billions of parameters (e.g., GPT-3 has 175B parameters), making it impractical to finetune the whole models to fit RS data. Therefore, there calls for efficient model finetuning techniques that can adapt LLMs (such as LLaMA~\cite{touvron2023llama}) for RS image analysis tasks. There are three potential solutions: 1) prompt fine-tuning~\cite{brown2020language} that designs learnable prompts that are finetuned in new domains; 2) adapter networks~\cite{houlsby2019parameter,lin2020exploring} that insert adapter layers between existing layers in deep neural networks; 3) low-rank adaption that injects trainable rank decomposition matrices into each layer of Transformer architectures. For instance, recently proposed LoRA~\cite{hu2021lora} can reduce the number of trainable parameters by 10,000 times and reduce the GPU memory by three times, which can be a promising solution for finetuning large models on RS tasks.

    \item \textbf{Integrate RS expert knowledge into LLMs.} To better utilize LLMs for RS data analysis, an important step is to integrate RS expert knowledge into LLMs properly—this calls for empowering large language models with domain-specific knowledge about RS images, such as sensor imaging theory, spatial correlation, and spectral characteristics of ground objects. Recent work developed a new technique, called instruction tuning~\cite{wei2021finetuned}, to enhance the performance of LLMs under instructions. The instruction is finetuned on several full-shot tasks and then evaluated for its zero-shot generalization ability on specific tasks. In remote sensing, applying instruction fine-tuning can potentially enable knowledge-based instructions to generate and understand RS images.

    \item \textbf{Linking text-based information with RS via geolocation.} LLMs can be exploited to analyze text data associated with geolocation such as social media text messages \cite{zhu2022tweet}, newspapers, etc., to extract linguistic features or even geoinformation, which can then be further fused with remote sensing data. This opens up new perspectives for a wide range of applications, such as semantic understanding of buildings ~\cite{HABERLE2022255}, disaster response ~\cite{kruspe2021detection}, and geo-aware social dynamics~\cite{kruspe2020cross,rode2022true}, and offer new possibilities to utilize unconventional geodata sources that are complementary to remote sensing data. 

    \item \textbf{Climate Change Adaptation and Mitigation} Vision-Language Models (VLMs) present a comprehensive approach to support and enhance climate change adaptation and mitigation initiatives, integral to the United Nations Sustainable Development Goals (SDGs). For instance, VLMs can play a vital role in monitoring changes in forest cover and bolstering forest conservation efforts. VLMs possess the ability to seamlessly integrate satellite imagery with forest-related textual data and scientific reports. This integration facilitates the automatic recognition and categorization of critical forest landmarks and diverse vegetation species, thereby enabling real-time monitoring of changes in forest cover, fire-prone areas, and vegetation transitions. By utilizing VLMs, we can effectively curb greenhouse gas emissions, maintain ecological equilibrium, and address the challenges posed by climate change in a proactive manner.
    
\end{itemize}

% \section{Author Contribution}
% Xiang Li conceived and designed the study, and wrote sections I, II.B, III.E, III.G, III.H, III.I, and IV. Congcong Wen wrote sections II.C, III.B, III.C, III.D, and IV. Yuan Hu wrote sections II.A, III.A, III.H, and III.I. Zhenghang Yuan and Xiao Xiang Zhu wrote section III.F.

\bibliography{egbib}
\bibliographystyle{ieee}

\end{document}